\documentclass[journal,twoside,web]{ieeecolor}
\usepackage{generic}
\usepackage{cite}
\usepackage{amsmath,amssymb,amsfonts}
\usepackage{algorithmic}
\usepackage{graphicx}
\usepackage{algorithm,algorithmic}
\usepackage{hyperref}
\usepackage{multirow}
\usepackage{multicol}
\hypersetup{hidelinks=true}
\usepackage{textcomp}
\usepackage[final]{changes}
\def\BibTeX{{\rm B\kern-.05em{\sc i\kern-.025em b}\kern-.08em
    T\kern-.1667em\lower.7ex\hbox{E}\kern-.125emX}}
\begin{document}

\title{Estimation of Segmental Longitudinal Strain in Transesophageal Echocardiography by Deep Learning}

\author{Anders Austlid Taskén, Thierry Judge, Erik Andreas Rye Berg, Jinyang Yu, Bjørnar Grenne, Frank Lindseth, Svend Aakhus, Pierre-Marc Jodoin, Nicolas Duchateau, Olivier Bernard, and Gabriel Kiss
\thanks{Submitted on \today. This work was supported through internal funding from The Department of Computer Science (IDI), Norwegian University of Science and Technology (NTNU) and the French National Research Agency (ANR) through the ORCHID [ANR-22-CE45-0029-01] project.}  
\thanks{Anders Austlid Taskén, Frank Lindseth and Gabriel Kiss are with the Department of Computer Science, Faculty of Information Technology and Electrical Engineering, Norwegian University of Science and Technology, 7034 Trondheim, Norway (e-mail: anders.a.tasken@ntnu.no; frankl@ntnu.no; gabriel.kiss@ntnu.no).}
\thanks{Thierry Judge and Pierre-Marc Jodoin are with the Department of Computer Science, University of Sherbrooke, Sherbrooke, QC J1K 2R1, Canada.}
\thanks{Erik Andreas Rye Berg, Jinyang Yu, Bjørnar Grenne and Svend Aakhus are with the Department of Circulation and Medical Imaging, Faculty of Medicine and Health Sciences, Norwegian University of Science and Technology, 7030 Trondheim, Norway.}
\thanks{Erik Andreas Rye Berg, Bjørnar Grenne and Svend Aakhus are with the Clinic of Cardiology, St. Olavs University Hospital, Trondheim, Norway.}
\thanks{Jinyang Yu is with the Clinic of Anesthesia and Intensive Care, St. Olavs University Hospital, Trondheim, Norway.}
\thanks{Olivier Bernard and Nicolas Duchateau are with the INSA-Lyon, Universite Claude Bernard Lyon 1, CNRS, Inserm, CREATIS UMR 5220, U1294, 69621 Lyon, France.}
\thanks{Nicolas Duchateau is also with the Institut Universitaire de France (IUF).}}

\maketitle

\begin{abstract}
Segmental longitudinal strain (SLS) of the left ventricle (LV) is an important prognostic indicator for evaluating regional LV dysfunction, in particular for diagnosing and managing myocardial ischemia. Current techniques for strain estimation require significant manual intervention and expertise, limiting their efficiency and making them too resource-intensive for monitoring purposes. This study introduces the first automated pipeline, autoStrain, for SLS estimation in transesophageal echocardiography (TEE) using deep learning (DL) methods for motion estimation. We present a comparative analysis of two DL approaches: TeeFlow, based on the RAFT optical flow model for dense frame-to-frame predictions, and TeeTracker, based on the CoTracker point trajectory model for sparse long-sequence predictions.

As ground truth motion data from real echocardiographic sequences are hardly accessible, we took advantage of a unique simulation pipeline (SIMUS) to generate a highly realistic synthetic TEE (synTEE) dataset of 80 patients with ground truth myocardial motion to train and evaluate both models. Our evaluation shows that TeeTracker outperforms TeeFlow in accuracy, achieving a mean distance error in motion estimation of 0.65 $\pm$ 0.20 mm on a synTEE test dataset. 

Clinical validation on 16 patients further demonstrated that SLS estimation with our autoStrain pipeline aligned with clinical references, achieving a mean difference (95\% limits of agreement) of 1.09\% (-8.90\% to 11.09\%). Incorporation of simulated ischemia in the synTEE data improved the accuracy of the models in quantifying abnormal deformation. Our findings indicate that integrating AI-driven motion estimation with TEE can significantly enhance the precision and efficiency of cardiac function assessment in clinical settings.

\end{abstract}


\section{Introduction}
\label{sec:introduction}

\IEEEPARstart{I}{n} the diagnosis and treatment of perioperative patients with wall motion abnormalities, such as myocardial ischemia, assessment of regional left ventricular (LV) systolic function is valuable \cite{perioperative_sls}. Segmental longitudinal strain (SLS) by echocardiography, describing the deformation of the myocardium, is a descriptive cardiac functional index that provides important information about mechanical impairment \cite{ste_bart}. \replaced{SLS measures the strain in individual myocardial segments, enabling the detection of regional wall motion abnormalities. In contrast, Global Longitudinal Strain (GLS) represents the average strain across the entire left ventricle, providing a single value that reflects overall LV function. Both SLS and GLS are critical for assessing cardiac health, with SLS being particularly sensitive to regional dysfunction \cite{sls_1, sls_2}. These metrics can be estimated from routine transesophageal echocardiography (TEE) sequences perioperatively.}{It is a sensitive marker of regional LV dysfunction \cite{sls_1, sls_2}, and can be estimated from routine transesophageal echocardiography (TEE) sequences perioperatively.} Robust assessment of regional LV function is imperative for early detection of myocardial ischemia, but currently available tools for quantitative assessment of SLS are subjective, require high expertise, are resource intensive, and exhibit limited reproducibility.

Currently, SLS is determined from echocardiographic image sequences using a conventional motion estimation technique known as speckle tracking echocardiography (STE) \cite{ste}. STE assesses the motion of pixel blocks within the myocardial wall. Although widely embraced, in particular in commercial software, STE encounters difficulties associated with the intrinsic characteristics of ultrasound imaging \cite{ste_commercial}. Challenges such as reverberation, shadows, out-of-plane motion, and dropouts contribute to substantial frame-to-frame decorrelation of the speckle pattern, potentially leading to poor tracking performance. Furthermore, in clinical use, numerous steps of manual intervention are needed to initiate and correct STE. This includes adjustment of the boundaries of the myocardial wall, view selection, and tuning of algorithmic parameters. Therefore, STE is both time-consuming and requires substantial expertise. 

In the perioperative setting when the patients are mechanically ventilated, the use of TEE is preferable to standard transthoracic echocardiography (TTE) because it offers superior image clarity \cite{VIGNON19941829, GARCIA2017736}. Additionally, employing TEE allows the transducer to remain passive in the esophagus, which not only allows continuous monitoring but also minimizes esophageal trauma \cite{eab_automapse, yu, yu_2}. \added{TEE monitoring with passive probe placement presents specific challenges distinct from TTE, including increased foreshortening, different speckle patterns from esophageal positioning, and unique noise characteristics from prolonged esophageal contact.} However, most methods available for motion estimation in echocardiography are optimized for TTE. Our study was targeted on TEE data. 

Recent advances in artificial intelligence (AI) have demonstrated encouraging outcomes in the realm of optical flow estimation and tracking-based motion estimation. State-of-the-art (SOTA) methods for motion estimation by deep learning (DL) in echocardiography have predominantly used dense frame-to-frame motion estimation, thus only incorporating information between two consecutive images. Among these methods, the RAFT model \cite{raft} currently represents the most successful approach, achieving significant improvements in cases of fast movement and motion blur. However, optical flow-based methods often struggle with occlusion and drift over long sequences, limiting their performance in real-world applications. Tracking-based methods, such as CoTracker \cite{cotracker}, estimate motion by tracking sparse points over extended periods, improving accuracy and robustness in handling occlusions and complex motion patterns.

In this paper, we present the first study to compare the performance of two distinct SOTA DL approaches for the estimation of myocardial motion in transesophageal echocardiography (TEE): 1) TeeFlow, a RAFT-based neural network for optical flow estimation, and 2) TeeTracker, a CoTracker-based neural network for point trajectory estimation. TeeFlow estimates myocardial motion by predicting dense displacement fields between consecutive frames, while TeeTracker estimates point trajectories over an entire sequence, directly providing myocardial mesh positions throughout the cardiac cycle. 

The training and validation of motion estimation methods is challenging due to the need of echocardiographic sequences with ground truth motion references. Such data are rarely accessible and are nonexistent within the domain of TEE data. Thus, we present the first publicly available dataset of synthetic TEE data with ground truth references of myocardial motion, referred to as synTEE. We adopted a simulation pipeline (SIMUS) to generate the synTEE data, which enabled the training and testing of TeeFlow and TeeTracker \cite{creatis_synTTE_strain}. Several simulation strategies with various amounts of decorrelation were applied during training and evaluation in order to produce simple to hard tracking scenarios to facilitate the applicability on real-world data. In addition, we achieved greater variability of deformation patterns in the synthetic dataset by simulating regional wall abnormalities. More specifically, we artificially decreased myocardial deformation in specific segments to mimic myocardial infarction.

\subsection{Previous Work}
\label{sec:previous_work}

\subsubsection{Motion Estimation}

DL-based methods have dominated the field of motion estimation in the last decade. FlowNet showed the potential of DL by directly estimating the optical flow from consecutive images \cite{flownet}, and FlowNet2 outperformed previous classical approaches to general optical flow problems \cite{flownet2}. PWC-Net decreased the speed of inference while increasing performance by combining a pyramidal structure with warping and cost volumes \cite{pwcnet}. RAFT introduced a recurrent neural network for iteratively refining the motion estimation, and improved the performance in cases of fast movement and motion blur \cite{raft}. FlowFormer proposed a transformer-based method to learn a 4D cost volume from image pairs, similar to RAFT. This approach involved tokenizing the cost volume and encoding it into a cost memory using a transformer. Subsequently, a recurrent transformer was employed to decode the cost memory, resulting in a network that outperformed the SOTA methods on the benchmarks \cite{flowformer}. The optical flow based methods struggled with occlusion and drifting, since the motion was only predicted between consecutive frames. Tracking-based methods recently entered the stage, performing motion estimation on a sparse set of points over an extended period of time. TAP-Vid \cite{tapnet} and PIPs \cite{pips} demonstrated the potential of DL in tracking, and TAPIR \cite{tapir} and PIPs++ \cite{pips++} successfully tracked occluded points on synthetic benchmarks. CoTracker further improved tracking accuracy and robustness by jointly tracking dense points across the video, in contrast to previous methods treating points independently \cite{cotracker}. 

\subsubsection{Strain Estimation in Echocardiography}

Optical flow, as a method for myocardial motion estimation, has demonstrated its efficiency in capturing global contraction patterns. Østvik et al. adopted PWC-Net for TTE, named EchoPWC-Net, and trained it on a set of 105 simulated ultrasound sequences \cite{ostvik_strain}. Additionally, they integrated it into a pipeline designed for the estimation of global longitudinal strain (GLS), namely, strain averaged across the whole myocardium. Building on this work, Evain et al. introduced a novel simulation pipeline aimed at generating a more extensive synthetic dataset of TTE sequences \cite{creatis_synTTE_strain}. They then trained a modified version of PWC-Net specifically tailored for optical flow estimation in TTE, achieving comparable results in GLS estimation to those reported by Østvik et al. Deng et al. took a different approach by incorporating the RAFT model into a strain estimation pipeline \cite{raft_tte}. Their integration led to further enhancements in GLS estimation when compared against clinical references. Azad et al. proposed a novel method performing point trajectory estimation for GLS estimation, further improving the accuracy and precision of global LV estimation in TTE \cite{azad2024echo}. This method conducts a two-fold coarse-to-fine model to track myocardial points in TTE inspired by TAPIR \cite{tapir}. Contrary to this work that may introduce ambiguity because of intrinsic inter- and intra-expert variabilities, we focus on generating realistic simulations with different degrees of speckle decorrelation and myocardial infarct to train our DL solutions. 

\subsubsection{Echocardiography Simulation}

Successful estimation of myocardial motion through DL hinges on the availability of a high quality annotated dataset that contains ground truth references for myocardial motion patterns throughout the cardiac cycle. Given the substantial labor and clinical expertise required for manual annotation, the development of a pipeline to generate synthetic ultrasound sequences has emerged as a valuable approach to procuring sufficiently large datasets for model training and validation \cite{3d_synthetic_echo, ste_standardization}. Alessandrini et al. have contributed to this effort by introducing an openly accessible database that includes realistic vendor-specific synthetic ultrasound data, derived from biomechanically personalized simulations \cite{syntte_v1}. This database was originally intended to validate speckle tracking algorithms, a purpose for which Østvik et al. used it in the training of EchoPWC-Net. Subsequently, Evain et al. presented a dedicated simulation strategy capable of generating an even larger dataset with realistic tissue texture \cite{creatis_synTTE_strain}. Notably, despite these advancements, to date, no dataset containing ground truth myocardial motion data has been specifically tailored for TEE data. \added{While existing synthetic databases like Alessandrini et al. provide valuable resources for TTE, their B-mode warping approach enforces unrealistically high temporal speckle correlation - a limitation particularly problematic for TEE monitoring scenarios where probe movement and tissue dynamics create natural decorrelation. Our TEE-specific framework overcomes this by providing independent control over scatterer coherence ratios, enabling systematic investigation of decorrelation effects on tracking performance under clinically relevant conditions, including prolonged esophageal contact and LV foreshortening artifacts.}

\subsection{Main Contributions}

We propose a pipeline for motion estimation and SLS prediction in TEE. The main contributions of this paper are: 

\begin{itemize}
    \item Comparison of two 
    SOTA 
    DL strategies for myocardial motion estimation in TEE: dense motion estimation (TeeFlow) and point trajectory estimation (TeeTracker).
    \item Development of a novel pipeline for automatic estimation of regional LV function by predicting SLS 
    in critically ill patients.
    \item Creation of an open-access database with 240 synthetic TEE sequences from 80 patients, with ground truth references of myocardial motion and with various degrees of decorrelation and synthetic myocardial infarctions for realistic simulations.
    \item Setup and evaluation of a comprehensive and novel set of experiments to obtain the best motion estimation methods on real data.
\end{itemize}

By conducting this comparative study, our aim is to advance the field of echocardiography by identifying the most effective DL-based method for the estimation of SLS, ultimately improving the perioperative monitoring and treatment of critically ill patients.

\section{Materials and Method}

\subsection{Data Acquisition and Study Subjects}

2D TEE B-mode ultrasound images of a total of 80 patients were anonymized and included in the study. The ultrasound probe was unlocked without forced flex or tilt in all recordings included in this prospective study, to assess the feasibility of cardiac monitoring. Thus, data were acquired without direct interference by a physician by leaving the probe passively in the esophagus. This passive placement of the probe resulted in foreshortened images with increased noise and out-of-plane movement, producing sequences with potentially high frame-to-frame decorrelation. Such a scenario makes tracking more challenging. Esophageal trauma was minimized due to an unlocked probe tip, which was common practice during surgery. 

Cardiologists with expertise in echocardiography collected the data at the Echolab, St. Olavs University Hospital in Trondheim, Norway. A GE Vivid E95 scanner with a 6VT-D probe was used to acquire the images (GE Vingmed Ultrasound, Horten, Norway). Patients were excluded if they were below 18 years old, or had clinical contraindications to TEE. All patients were provided with written details about the study and had the option to withdraw. The study did not interfere with standard clinical care. This study was approved by the Regional Committees for Medical and Health Research Ethics (REK 2017/900). 

The study included patients undergoing routine care in the echocardiography unit and were related to interventional cardiology, electrophysiology, and cardiothoracic surgery. Mid-esophageal 4-chamber (4C), 2-chamber (2C) and long-axis (LAX) views of the LV were acquired separately and sequentially. Furthermore, all ultrasound images were acquired with a fixed sector width of 90 degrees and a depth adjusted to include the apex of the LV. 

\subsection{Data Annotation}

\replaced{The endocardium and epicardium borders were manually annotated at a mid-systolic frame presented through the commercial software EchoPAC (version 204, GE, Vingmed Ultrasound, Horten, Norway) and tracked through the cardiac cycle using speckle tracking technology with expert corrections when necessary.}{The endocardium and epicardium borders were manually annotated in a single frame and tracked through the cardiac cycle by speckle tracking technology (STE) and corrected by a cardiologist with expertise in echocardiography, in the case of need.These clinical references were obtained with the 2D Strain from a commercial software (EchoPAC version 206, GE, Vingmed Ultrasound, Horten, Norway)}. The annotations were exported with internal software provided by GE Vingmed. \added{All acquired segments were analyzed without exclusions, including those with suboptimal image quality, to provide a complete assessment of method performance under uncurated clinical conditions.}

\subsection{In Silico Modeling of Myocardial Motion in TEE}

\begin{figure*}[tbph]
\centerline{\includegraphics[width=\textwidth]{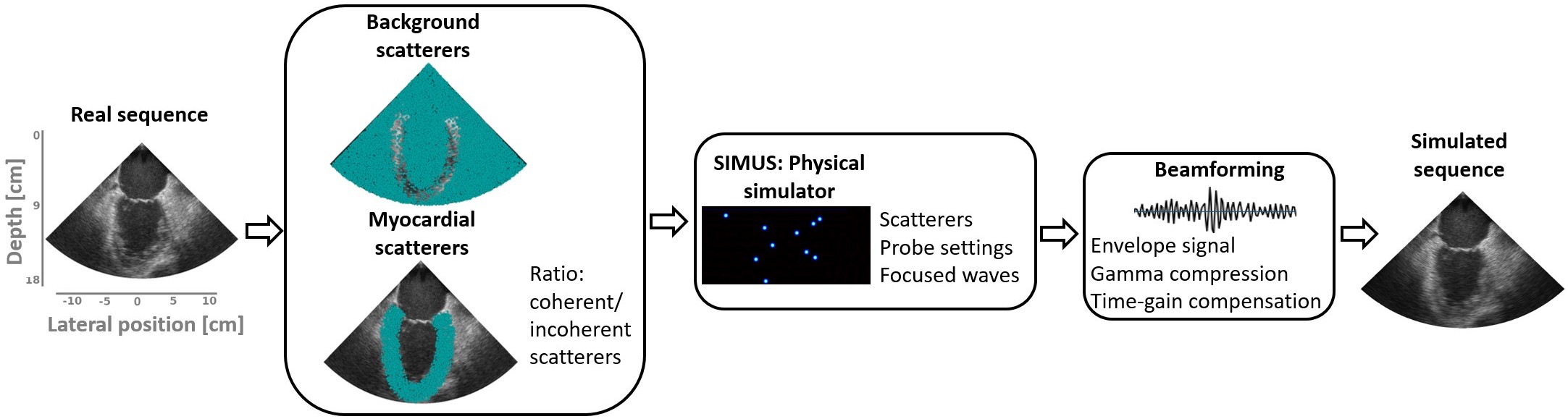}}
\caption{Simulation pipeline that was deployed to generate synthetic 2D TEE B-mode sequences with reference myocardial contraction fields, referred to as the synTEE dataset.}
\label{fig:simulation_pipeline}
\end{figure*}

For supervised learning of motion estimation using DL, a dataset with ground truth references was needed for the myocardial motion field. Collecting this information through manual annotation is a laborious task that limits the size of the corresponding dataset and introduces inter-/intraobserver variability that can confuse the algorithm during the learning phase. Thus, we decided to simulate the first synthetic 2D TEE B-mode sequences with reference myocardial contraction fields using the same formalism as the one developed by Evain et al. \cite{creatis_synTTE_strain}. We simulated 4C, 2C and LAX views of the LV. The simulation pipeline used a sequence of real TEE frames to first estimate relevant scatter maps that were used as references for the myocardial motion. These scatterers were then fed into a physical simulator (SIMUS software \cite{simus}) to generate raw data, which were finally post-processed using a conventional beamforming technique to obtain realistic TEE simulated sequences. Figure \ref{fig:simulation_pipeline} gives an overview of the full simulation pipeline that we used. The complete dataset has been released for access by researchers \cite{syntee_public}.

\subsubsection{Simulation of Decorrelation}
\label{sec:simulation_decorrelation}

We adapted the simulation pipeline to generate four different synthetic sequences for each patient. These sequences share the same myocardial motion, but have different speckle patterns, with varying degrees of decorrelation across the myocardial region. This was accomplished by adjusting the ratio between coherent and incoherent scatterers within the myocardium. In their seminal paper \cite{creatis_synTTE_strain}, the authors set this ratio at 0.9 (i.e., 90\% coherent scatterers and 10\% incoherent scatterers) to maintain high correlation within the myocardium throughout the cardiac cycle. In this study, we propose to adjust this ratio to control the degree of speckle decorrelation in the myocardium, generating TEE B-mode synthetic sequences with varying levels of difficulty. These ratios were chosen empirically and the corresponding values were reported in table \ref{tab:datasets_ratio}. 

We chose to focus on speckle decorrelation for several reasons. First, it is well known that speckle decorrelation naturally varies along the myocardium over time due to the intrinsic nature of echocardiography. Simulating sequences with overly high correlation therefore reduces the realism of the simulated data. Additionally, since we worked with unlocked transesophageal probes, LV foreshortening phenomena could occur, promoting out-of-plane movements that resulted in increased speckle decorrelation in the images. By integrating speckle decorrelation into our simulation pipeline, we could indirectly account for these phenomena without explicitly modeling them.

\begin{table}[tbph]
\centering
\caption{Overview of synTEE datasets}
\label{tab:datasets_ratio}
\begin{center}
\resizebox{.8\columnwidth}{!}{%
\begin{tabular}{llc}
\hline
Dataset & Characteristics & Ratio \\ 
\hline
1       & Few decorrelation & 0.9 \\
2       & Mild decorrelation  & 0.7 \\
3       & Moderate decorrelation  & 0.6  \\
4       & Severe decorrelation & 0.5
\end{tabular}
}%
\end{center}
\hfill \break
{The properties of the four synthetic TEE datasets derived from different ratios of coherent to incoherent scatterers.}
\end{table}

\subsubsection{Simulation of Synthetic Infarction}

To account for patients with hypokinetic pathologies, we enhanced the simulation pipeline by designing additional scenarios that simulate myocardial infarction in one of the six cardiac segments. Based on the pipeline described above (c.f. Figure~\ref{fig:simulation_pipeline}), a synthetic myocardial motion was first estimated from a real sequence using the same technique as previously described \cite{creatis_synTTE_strain}. The longitudinal contraction of myocardial scatterers was then reduced locally (i.e. at the center of a specific segment) following a Gaussian distribution throughout the cardiac cycle. The scatterers in the surrounding tissues were finally designed to compensate for the reduced contractility in order to maintain overall contraction, thus enabling the rest of the pipeline to be preserved. 

\begin{figure*}[tbph] 
    \centering
    \begin{minipage}{.24\textwidth}
        \centering
        \includegraphics[width=\textwidth]{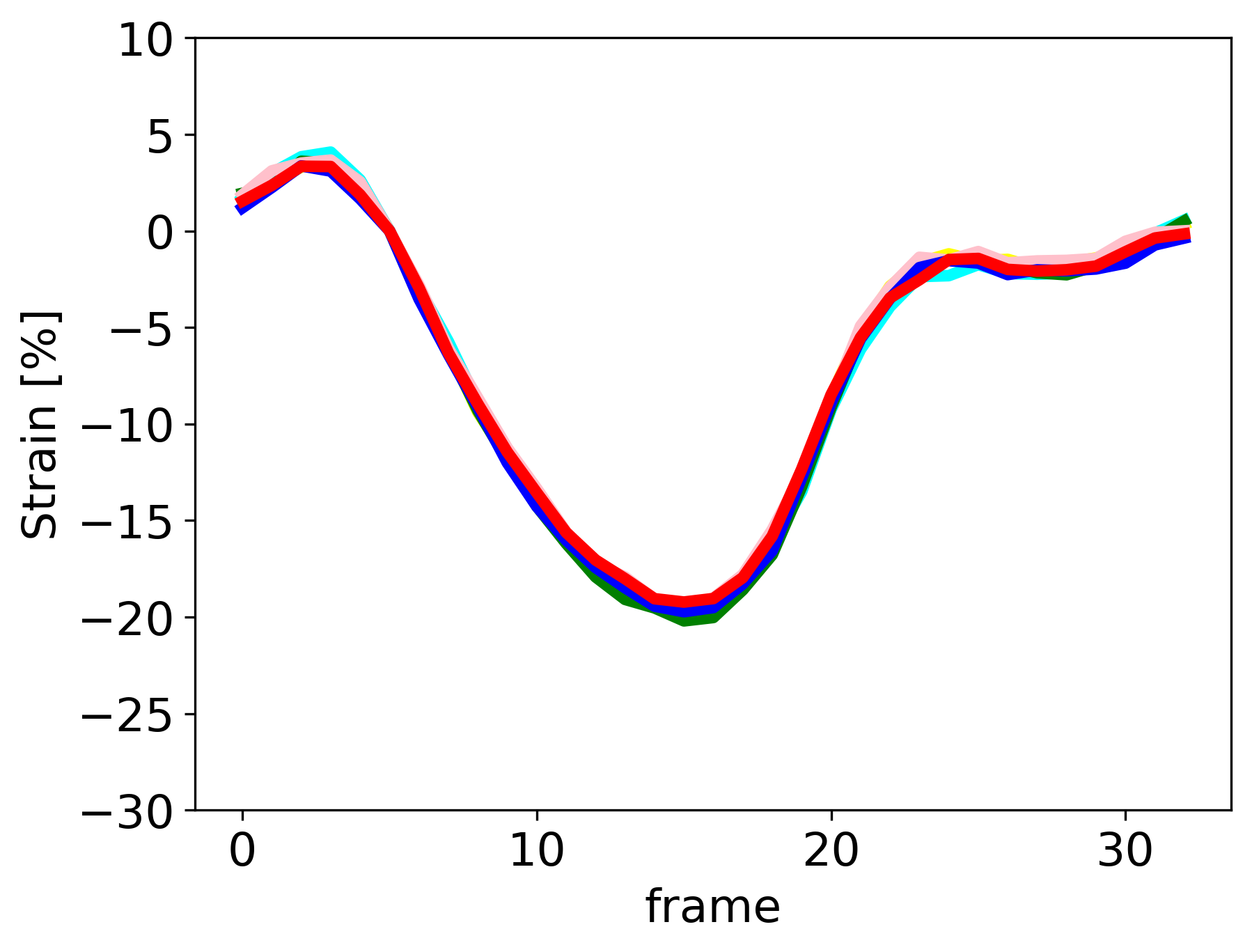}
        (a)
    \end{minipage}
    \begin{minipage}{.24\textwidth}
        \centering
        \includegraphics[width=\textwidth]{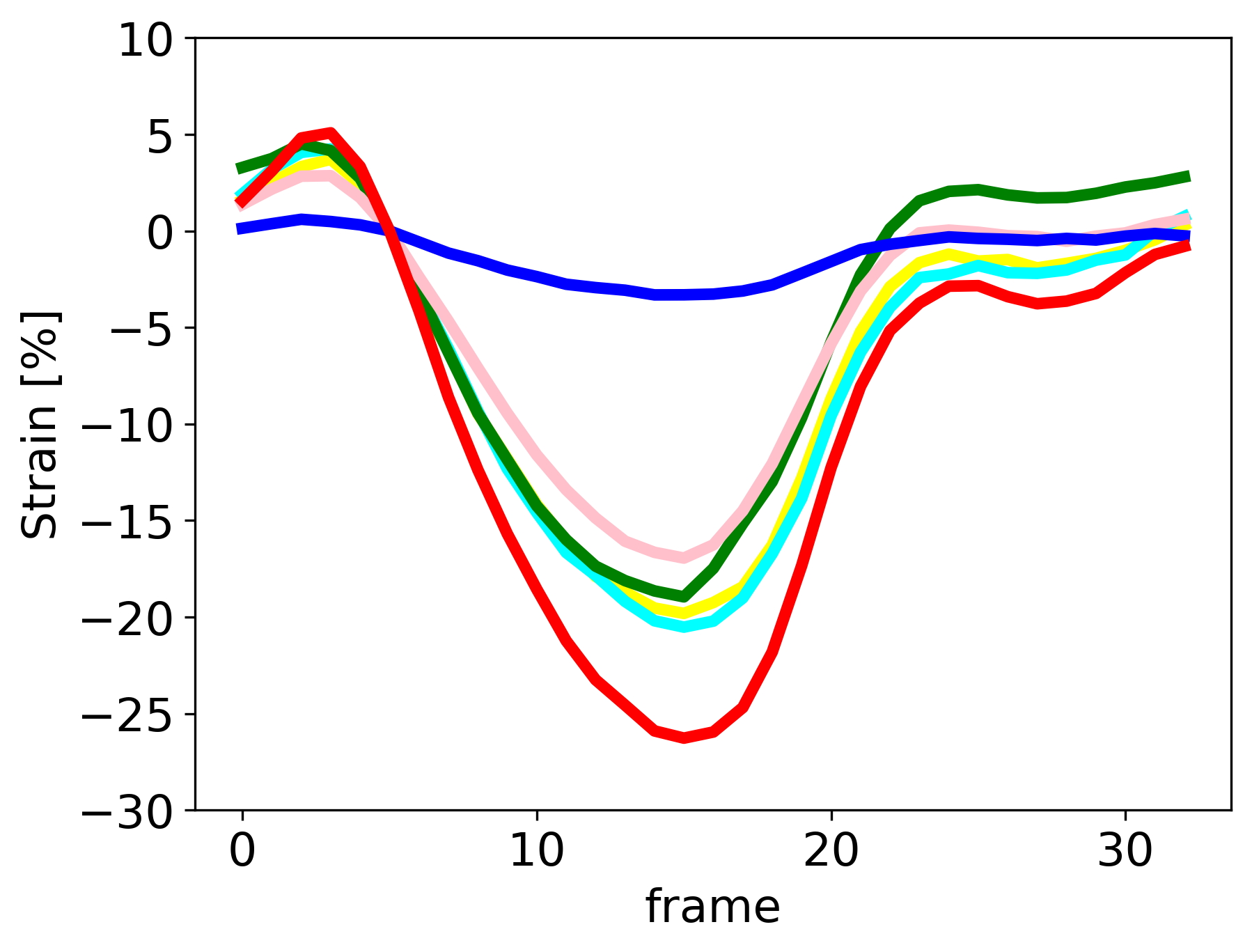}
        (b)
    \end{minipage}
    \begin{minipage}{.24\textwidth}
        \centering
        \includegraphics[width=\textwidth]{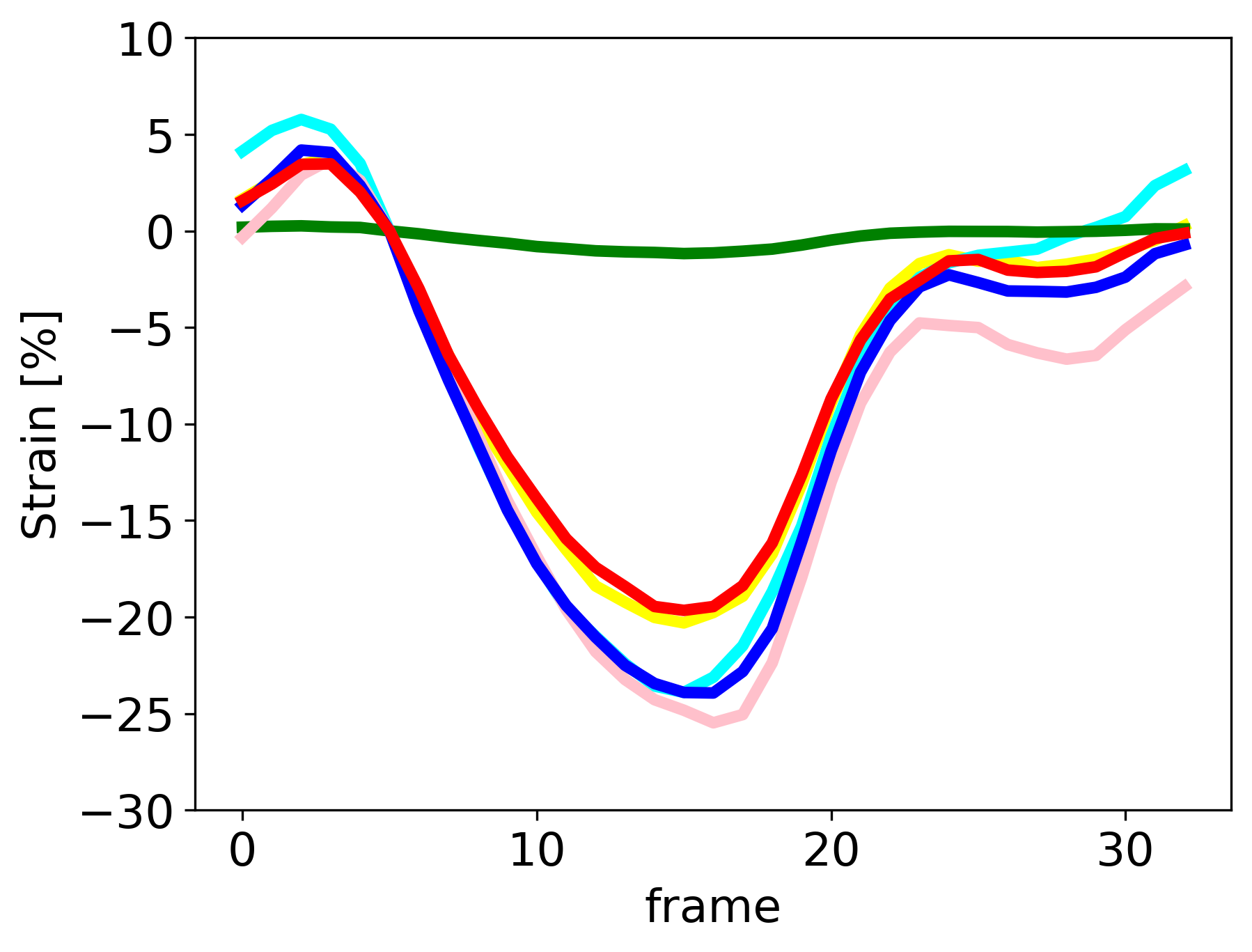}
        (c)
    \end{minipage}
    \begin{minipage}{.24\textwidth}
        \centering
        \includegraphics[width=\textwidth]{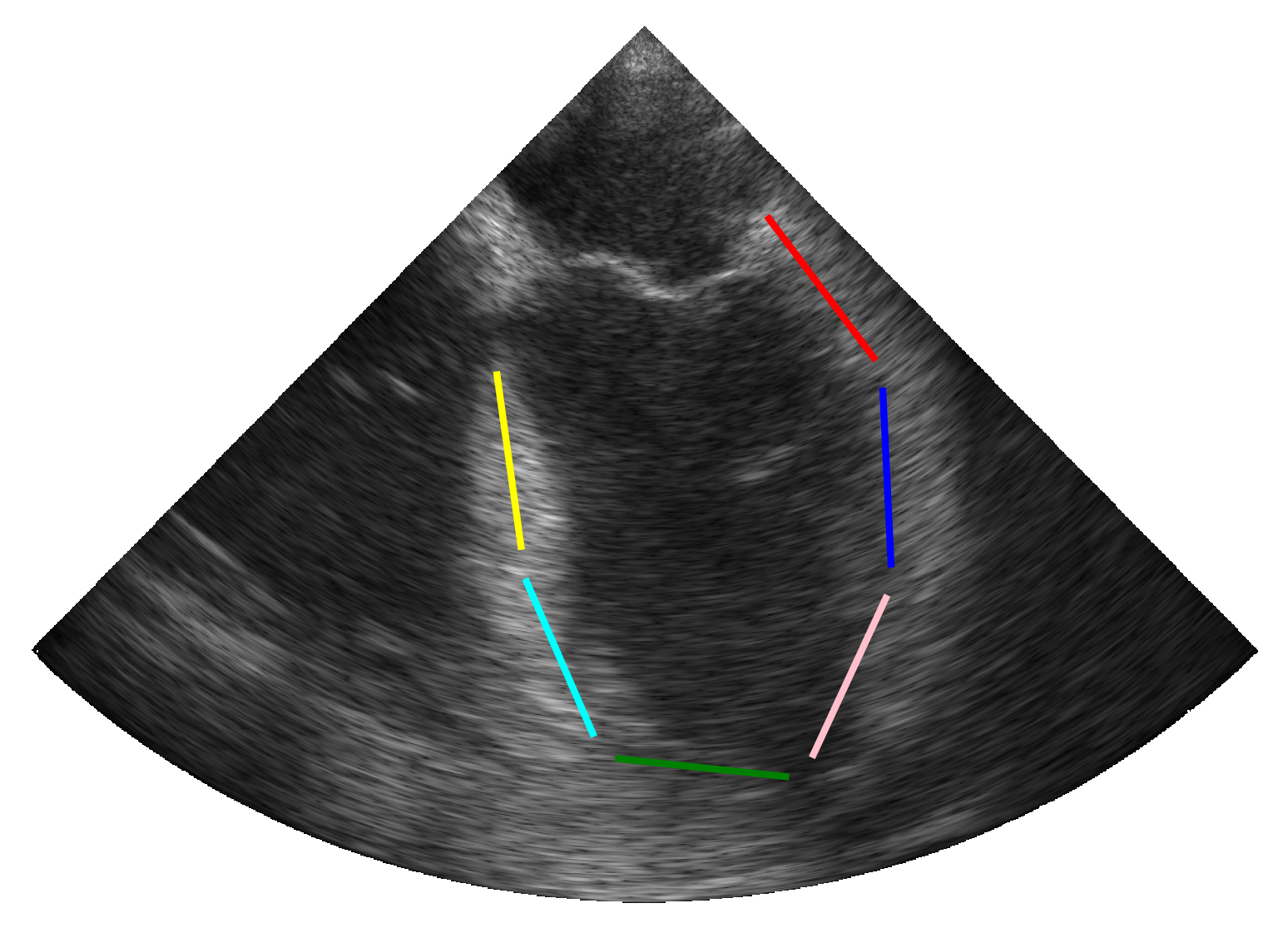}
        (d)
    \end{minipage}
    \caption{Examples of SLS curves after simulation of myocardial infarction, obtained in the synTEE dataset.  (a) original strain curves before manipulation; (b) and (c) reduced local myocardial deformation in 
    a given 
    segment; (d) visualization of the color map for myocardial segments.}
    \label{fig:prediction_examples}
\end{figure*}

\subsection{Myocardial Motion Estimation by Optical Flow}

Our TeeFlow algorithm was built upon the SOTA RAFT method \cite{raft} to achieve precise myocardial motion estimation in TEE sequences. RAFT is one of the most effective deep learning methods for estimating dense displacement fields representing pixel movements between successive images, making it a prime candidate for our task of estimating local myocardial deformation. This method is based on feature extraction blocks, correlation volumes, and an iterative update block that refines the estimated motion. To account for various motion amplitudes, multiple correlation matrices are computed at different scales and subsequently concatenated into a correlation pyramid. The displacement field is estimated iteratively using two convolutional gated recurrent unit (GRU) cells, simulating an optimization process. The GRU cell considers correlation features, current flow estimates, hidden features, and a context feature derived from the initial input image. After a predetermined number of iterative update steps, the final estimated displacement field is upsampled using a convex upsampling scheme to preserve the spatial resolution of the original data.

In the context of echocardiography imaging, feature encoders extract pertinent information from the input TEE images and visual similarity was quantified thanks to the correlation matrices able to handle the speckle patterns of ultrasound. The multiple correlation matrices accommodated for larger displacements typical in the basal segments of the LV. We carried out an extensive experimental plan to optimize this method for the unique characteristics of TEE imaging. We also conducted an ablation study to explore the benefits of fine-tuning the model parameters from pre-trained values optimized on synthetic natural scene sequences.

\subsection{Myocardial Tracking by Point Trajectory Estimation}

\begin{figure}[tbph]
\centerline{\includegraphics[width=\columnwidth]{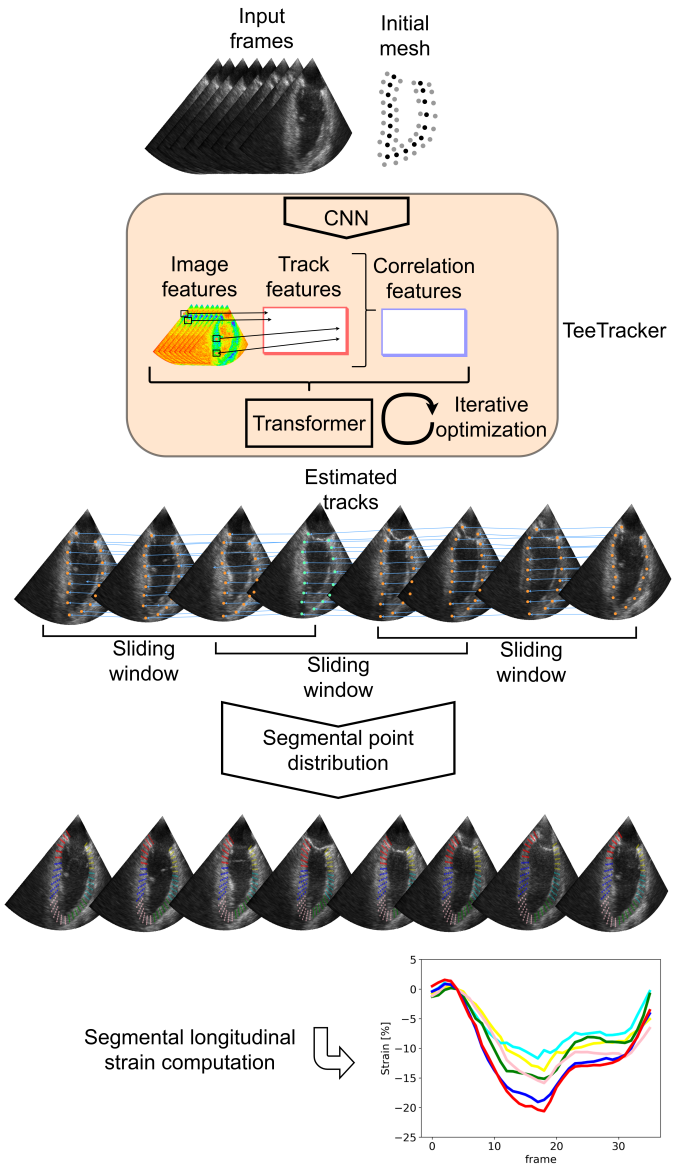}}
\caption{Visualization of myocardial tracking by point trajectory estimation with TeeTracker. TeeTracker had a sliding window length of 8 frames.}
\label{fig:TeeTracker_pipeline}
\end{figure}

TeeTracker was based on the CoTracker network \cite{cotracker} to obtain a precise estimation of the trajectories of the myocardial mesh in TEE (Figure \ref{fig:TeeTracker_pipeline}). CoTracker is a transformer-based network and a SOTA pipeline for point trajectory estimation, allowing tracking of sparse set of points in a video sequence. The network takes a set of consecutive images as input, collectively referred to as the `window', alongside a mesh representation of points of interest at a specified frame. Employing a CNN, CoTracker extracts critical features from the video sequence. These features forms the basis of input tokens, which encompasses image features, correlation vectors, visibility, appearance, and positional encodings. Subsequently, a transformed network iteratively processes these input tokens to refine the accuracy of track estimates, with initial track estimates assuming no motion. 

In the context of echocardiography, TeeTracker capitalized on the specific characteristics of ultrasound data, such as the spatial and temporal coherence of speckle patterns, to achieve a detailed estimation of myocardial motion dynamics. TeeTracker had a fix-sized sliding window to process the full TEE sequence at inference. To optimize the initialization of TeeTracker, we extracted the features at the end-systole (ES) frame such that the image-, track- and appearance features contained a minimum amount of noise, and such that most of the myocardium was visible. From the ES frame, we tracked the myocardial mesh both backward and forward to obtain an estimate of the point trajectories for the full sequence, leveraging the temporal coherence and continuity of ultrasound data. This non-online approach was feasible for this study and allows for comprehensive trajectory estimation, addressing the challenges posed by TEE.

\subsection{Training Procedures}
\label{sec:training_procedures}

Based on the four synthetic datasets (Table \ref{tab:datasets_ratio}), we explored various training schemes to fully leverage the varying degrees of speckle decorrelation for an efficient estimate of myocardial motion in TEE sequences. To evaluate the value of our synthetic dataset, we first conducted two ablation studies where the proposed algorithms were tested using their pre-trained parameters without any fine-tuning on the simulated TEE sequences. The corresponding results were given through the {\em pre-trained} lines in Table \ref{tab:synTEE_results}.

Each simulated dataset was then used individually to fine-tune TeeFlow and TeeTracker. Dataset 1 corresponds to the synthetic sequences with minimal decorrelation. It enables learning fundamental patterns of myocardial contraction without the confounding effects of noise. Datasets 2 and 3 gradually incorporate speckle decorrelation to challenge the models and enhance their robustness to typical in-plane motion variations and noise levels found in real data. Dataset 4 includes highly decorrelated sequences, simulating scenarios with significant noise and pronounced out-of-plane artifacts. This dataset was designed to increase the capacity of the networks to handle the most challenging and realistic conditions. The corresponding results were shown in rows labeled {\em dataset 1, 2, 3}, and {\em 4} (Table \ref{tab:synTEE_results}).

In addition, we also explored two advanced training strategies with the four datasets in a sequential or a combined scheme. In sequential training, the DL models were trained successively on the four datasets, starting with dataset 1, which had the lowest speckle decorrelation, and progressing through datasets 2, 3, and 4, each with increasing levels of difficulty. This progressive exposure to speckle decorrelation was designed to gradually acclimate the model to more realistic motion artifacts, potentially enhancing its adaptability and robustness. In the combined training, the models were trained directly from all 4 datasets. This allowed learning from a wide spectrum of decorrelation scenarios simultaneously. We hypothesize that such an approach forced the networks to handle the fundamental patterns of myocardial contraction while learning to handle various degrees of decorrelation.

\subsection{Pipeline for Segmental Longitudinal Strain Estimation}
\label{sec:autostrain}

The SLS estimation pipeline, dubbed {\em autoStrain}, was initiated by providing a myocardial mesh of a specified frame together with a full cycle acquisition of TEE, in addition to end-diastole (ED) and ES timings. The pipeline differed slightly for motion estimation using optical flow and for myocardial tracking using point trajectory estimation. In the case of optical flow, the mesh was updated by warping the positions based on the frame-to-frame estimated displacement field. In the case of point trajectory estimation, the mesh positions of the complete sequence were given directly by the model output. The change in longitudinal myocardial length was calculated per segment throughout the cardiac cycle and was given as a percentage of the longitudinal length in the ED. \replaced{For segmental analysis, we employed the AHA 18-segment model \cite{aha_18_seg}, dividing the LV into six segments at basal and mid-ventricular levels and four segments at the apical level, plus the apex itself.}{We used the 18-segment model to divide the myocardium \cite{aha_18_seg}.}

\subsection{Evaluation and Statistical Analysis}

To compare the performance of optical flow- and point-trajectory models, we computed the mean distance error between the estimated and reference meshes along the full sequence in millimeters. This metric was only computed on synthetic data due to the need for ground truth meshes. To validate the ability to estimate SLS on TEE, we compared the autoStrain estimations with clinical references. We evaluated the agreement on both synthetic and real TEE acquisitions. The analyses were completed as a blind study, without prior knowledge of the clinical measurements. 

We evaluated the agreement between manually and automatically estimated SLS and GLS measurements using Bland-Altman analysis. The results were reported as bias and 95\% limits of agreement (LoA) along with a Bland-Altman plot. Descriptive measures were reported as numbers or mean (standard deviation (SD)), unless otherwise specified. For continuous variables, we reported the mean ± standard deviation, while for dichotomous data, we present the results as numbers (percentages).

\subsection{Implementation Details}

Neural networks were trained using Pytorch and an Nvidia Quadro RTX 8000 GPU with 48GB GDDR6, which accelerated the training process. The test dataset was evaluated using an 11th Gen Intel® Core™ i7-11850H @ 2.50GHz CPU and an NVIDIA RTX A3000 Mobile GPU. \added{TeeFlow achieved a mean frame rate of 5.60 frames per second, while TeeTracker achieved a mean frame rate of 1.59 frames per second. Both methods significantly outperform manual analysis, which typically requires 1-5 minutes per cardiac cycle.} All networks were trained using a supervised learning approach with training and validation data. TeeFlow and TeeTracker were trained for 5 000 and 50 000 steps, respectively. TeeFlow was trained with a batch size of 16, while TeeTracker was trained with a batch size of 1. \added{For TeeTracker, we employed a composite loss function with sequence and visibility prediction components, using AdamW optimization with a one-cycle learning rate schedule over approximately 25 epochs.} An AdamW optimizer was employed for both models, with a learning rate of 0.0001 and 0.0005 for TeeFlow and TeeTracker, respectively. The state of the models at the epoch with the lowest validation loss was used to test the model performance.

\subsection{Implementation Details}
Neural networks were trained using Pytorch and an Nvidia Quadro RTX 8000 GPU with 48GB GDDR6, which accelerated the training process. The test dataset was evaluated using an 11th Gen Intel® Core™ i7-11850H @ 2.50GHz CPU and an NVIDIA RTX A3000 Mobile GPU. \added{TeeFlow achieved a mean frame rate of 5.60 frames per second, while TeeTracker achieved a mean frame rate of 1.59 frames per second. Both methods significantly outperform manual analysis, which typically requires 1-5 minutes per cardiac cycle.} All networks were trained using a supervised learning approach with training and validation data. TeeFlow and TeeTracker were trained for 5 000 and 50 000 steps, respectively. TeeFlow was trained with a batch size of 16, while TeeTracker was trained with a batch size of 1. **For TeeTracker, we employed a composite loss function with sequence and visibility prediction components, using AdamW optimization with a one-cycle learning rate schedule over approximately 25 epochs.** An AdamW optimizer was employed for both models, with a learning rate of 0.0001 and 0.0005 for TeeFlow and TeeTracker, respectively. The state of the models at the epoch with the lowest validation loss was used to test the model performance.

\section{Results}

\subsection{Synthetic Data}


\subsubsection{Influence of the Training Procedure}

Table \ref{tab:synTEE_results} provides the geometric accuracy scores of TeeFlow and TeeTracker when evaluated on synthetic datasets employing the protocols procedures detailed in Section \ref{sec:training_procedures}. The table highlights the significant reduction in mean distance error following fine-tuning, compared to the pre-trained versions of the models. Specifically, the mean distance error for the best TeeFlow model decreased from 3.28 mm to 1.55 mm, and for the best TeeTracker model, it decreased from 1.08 mm to 0.65 mm. These findings validated the efficacy of our fine-tuning approach, suggesting that the models were well adapted to real B-mode TEE sequences due to the highly realistic simulations.

Furthermore, Table \ref{tab:synTEE_results} reveals the superiority of TeeTracker over TeeFlow, regardless of the training scheme. Notably, the pre-trained TeeTracker (with a mean distance error of 1.08 mm) surpassed the best fine-tuned version of TeeFlow (with a mean distance error of 1.55 mm), demonstrating the superiority of point trajectory estimation over dense motion estimation in echocardiography. To our knowledge, this was the first time this finding has been clearly analyzed.

In-depth analysis of TeeTracker's performance reveals that when fine-tuned on synthetic sequences with minimal speckle decorrelation (dataset 1), the model achieved high accuracy in capturing fundamental myocardial contraction patterns in similar data (test set 1), with a mean distance error of 0.36 $\pm$ 0.06 mm. However, this training scheme showed limited generalizability to data with increased speckle decorrelation, with mean distance errors increasing to approximately 0.43 mm for datasets 2 and 3, and to 1.60 mm on dataset 4. Fine-tuning on dataset 2 or 3 led to slight improvements in those datasets (reducing the error from 0.43 mm to 0.40 mm) while maintaining accuracy on data with lower speckle decorrelation (test set 1). Despite these improvements, these models still struggle to generalize on synTEE data with severe decorrelation (1.60 mm on test set 4). Fine-tuning on highly decorrelated sequences (dataset 4) improved performance on similar data (mean distance error of 1.29 $\pm$ 0.41 mm). Unfortunately, this model failed to maintain good results on data with lower speckle decorrelation, with a mean distance error of 0.73 mm on test set 1 and 0.76 mm on test set 2 and 3. 

To capitalize on the insights provided by each synTEE dataset, two additional training procedures were investigated to leverage the entire synthetic dataset. The sequential procedure yielded results comparable to training solely on dataset 4. In contrast, the combined training approach produced the best overall performance, with an average mean distance error of 0.65 $\pm$ 0.20 mm calculated from the fourth test dataset. Based on these results, we decided to adopt the combined training scheme exclusively for the subsequent analyses in this article.

\begin{table*}[tbph]
\centering
\caption{Comparison of performance in synTEE with various training schemes}
\label{tab:synTEE_results}
\begin{center}
\resizebox{.9\textwidth}{!}{%
\begin{tabular}{llccllllllll}
\hline
\multicolumn{1}{c}{\multirow{3}{*}{Model}} & \multicolumn{1}{c}{\multirow{3}{*}{Training scheme}} & \multicolumn{10}{c}{Mean distance error {[}mm{]}}                                                                                                                                                                                                    \\ \cline{3-12} 
\multicolumn{1}{c}{}                       & \multicolumn{1}{c}{}                                 & \multicolumn{2}{l}{Test set 1}         & \multicolumn{2}{l}{Test set 2}                  & \multicolumn{2}{l}{Test set 3}                  & \multicolumn{2}{l}{Test set 4}                  & \multicolumn{2}{l}{Average}                     \\ \cline{3-12} 
\multicolumn{1}{c}{}                       & \multicolumn{1}{c}{}                                 & MD            & SD                     & \multicolumn{1}{c}{MD} & \multicolumn{1}{c}{SD} & \multicolumn{1}{c}{MD} & \multicolumn{1}{c}{SD} & \multicolumn{1}{c}{MD} & \multicolumn{1}{c}{SD} & \multicolumn{1}{c}{MD} & \multicolumn{1}{c}{SD} \\ \hline
\multirow{7}{*}{TeeFlow}                   & Pre-trained                                          & 3.15          & \textit{0.80}          & 3.27                   & \textit{0.89}          & 3.28                   & \textit{0.90}          & 3.42                   & \textit{0.93}          & 3.28                   & \textit{0.88}          \\
                                           & Dataset 1                                                    & 1.28          & \textit{0.42}          & 1.41                   & \textit{0.51}          & 1.43                   & \textit{0.54}          & 2.07                   & \textit{0.81}          & 1.55                   & \textit{0.57}          \\
                                           & Dataset 2                                                    & 1.32          & \textit{0.43}          & 1.39                   & \textit{0.49}          & 1.42                   & \textit{0.51}          & 2.05                   & \textit{0.85}          & 1.55                   & \textit{0.57}          \\
                                           & Dataset 3                                                    & 1.32          & \textit{0.43}          & 1.40                   & \textit{0.51}          & 1.42                   & \textit{0.51}          & 2.05                   & \textit{0.82}          & 1.55                   & \textit{0.57}          \\
                                           & Dataset 4                                                    & 1.43          & \textit{0.46}          & 1.55                   & \textit{0.54}          & 1.59                   & \textit{0.58}          & 2.00                   & \textit{0.73}          & 1.64                   & \textit{0.58}          \\
                                           & Sequential                                           & 1.54          & \textit{0.47}          & 1.60                   & \textit{0.53}          & 1.63                   & \textit{0.55}          & 1.95                   & \textit{0.73}          & 1.68                   & \textit{0.57}          \\
                                           & Combined                                             & 1.30          & \textit{0.42}          & 1.41                   & \textit{0.51}          & 1.44                   & \textit{0.53}          & 2.03                   & \textit{0.79}          & 1.55                   & \textit{0.56}          \\ \hline
\multirow{7}{*}{TeeTracker}                & Pre-trained                                          & 0.56          & \textit{0.16}          & 0.70                   & \textit{0.33}          & 0.69                   & \textit{0.32}          & 2.37                   & \textit{0.95}          & 1.08                   & \textit{0.44}          \\
                                           & Dataset 1                                                    & \textbf{0.36} & \textit{\textbf{0.06}} & 0.43                   & \textit{0.14}          & 0.42                   & \textit{0.13}          & 1.60                   & \textit{0.53}          & 0.70                   & \textit{0.22}          \\
                                           & Dataset 2                                                    & 0.35          & \textit{0.06}          & \textbf{0.40}          & \textit{0.13}          & \textbf{0.40}          & \textit{\textbf{0.10}} & 1.60                   & \textit{0.60}          & 0.69                   & \textit{0.22}          \\
                                           & Dataset 3                                                    & 0.40          & \textit{0.08}          & 0.46                   & \textit{0.15}          & 0.45                   & \textit{0.13}          & 1.63                   & \textit{0.54}          & 0.74                   & \textit{0.23}          \\
                                           & Dataset 4                                                    & 0.73          & \textit{0.17}          & 0.76                   & \textit{0.24}          & 0.76                   & \textit{0.25}          & \textbf{1.29}          & \textit{\textbf{0.41}} & 0.89                   & \textit{0.27}          \\
                                           & Sequential                                           & 0.77          & \textit{0.20}          & 0.80                   & \textit{0.30}          & 0.81                   & \textit{0.30}          & 1.50                   & \textit{0.50}          & 0.97                   & \textit{0.33}          \\
                                           & Combined                                             & 0.38          & \textit{0.07}          & 0.43                   & \textit{\textbf{0.12}} & 0.42                   & \textit{0.11}          & 1.38                   & \textit{0.49}          & \textbf{0.65}          & \textit{\textbf{0.20}} \\ \hline
\end{tabular}
}%
\end{center}
\hfill \break
{Comparison of performance at motion estimation between SOTA methods in synTEE with various training schemes, evaluated on test sets drawn from the various synthetic datasets (Table \ref{tab:datasets_ratio}). Motion estimation was evaluated directly by comparing the mean Euclidean distance between the reference and myocardial mesh throughout the entire sequence.}
\end{table*}

\subsubsection{Estimation of Strain Measures}

Table \ref{tab:synTEE_autoStrain_results} shows the GLS and SLS clinical scores reached by TeeTracker trained on the combined dataset and tested on the syntTEE data. These values were obtained using the \emph{autoStrain} pipeline described in Section \ref{sec:autostrain}. Regarding the GLS metric, our model achieved a mean difference (95\% limits of agreement) of 2.78\% ($-1.63\%$ to 7.19\%) compared with the ground truth meshes and computed across all synTEE test data. These results were consistent with those obtained with the best performing DL methods evaluated on TTE synthetic images \cite{creatis_synTTE_strain}. As for SLS, our method achieved an overall mean difference (95\% limits of agreement) of \mbox{$-0.38\%$ ($-5.00\%$ to 4.25\%)}. It was interesting to note the increased performance of our model, with a mean difference (95\% limits of agreement) of $-0.22\%$ ($-3.53\%$ to 3.09\%), when evaluated only on the basal and mid segments of the myocardium, which were known to be the most relevant for analysis in TEE acquisition.


\begin{table*}[tbph]
\caption{Performance of autoStrain in synTEE}
\label{tab:synTEE_autoStrain_results}
\begin{center}
\resizebox{.5\textwidth}{!}{%
\begin{tabular}{llcccc}
\hline
\multirow{2}{*}{Test set} & \multicolumn{1}{c}{\multirow{2}{*}{Myocardial segments}} & \multicolumn{2}{c}{SLS {[}\%{]}}        & \multicolumn{2}{c}{GLS {[}\%{]}}       \\ \cline{3-6} 
                          & \multicolumn{1}{c}{}                                     & MD             & SD                     & MD            & SD                     \\ \hline
\multirow{2}{*}{1}        & All                                                      & -0.31          & \textit{1.99}          & 2.83          & \textit{2.09}          \\
                          & Basal + mid                                              & -0.10          & \textit{1.32}          & 2.83          & \textit{2.09}          \\ \hline
\multirow{2}{*}{2}        & All                                                      & -0.33          & \textit{2.20}          & 2.83          & \textit{2.10}          \\
                          & Basal + mid                                              & -0.18          & \textit{1.35}          & 2.83          & \textit{2.10}          \\ \hline
\multirow{2}{*}{3}        & All                                                      & -0.30          & \textit{2.13}          & 2.85          & \textit{2.08}          \\
                          & Basal + mid                                              & -0.21          & \textit{1.38}          & 2.85          & \textit{2.08}          \\ \hline
\multirow{2}{*}{4}        & All                                                      & -0.57          & \textit{3.12}          & 2.59          & \textit{2.73}          \\
                          & Basal + mid                                              & -0.40          & \textit{2.69}          & 2.59          & \textit{2.73}          \\ \hline
\multirow{2}{*}{1+2+3+4}  & All                                                      & -0.38          & \textit{2.36}          & 2.78          & \textit{2.25}          \\
                          & \textbf{Basal + mid}                                     & \textbf{-0.22} & \textit{\textbf{1.69}} & \textbf{2.78} & \textit{\textbf{2.25}} \\ \hline
\end{tabular}
}%
\end{center}
\hfill \break
{Agreement of measures between estimated and reference SLS and GLS measures in synTEE. SLS and GLS were computed based on ground truth myocardial mesh. SLS, segmental longitudinal strain. GLS, global longitudinal strain. MD, mean difference. SD, standard deviation.}
\end{table*}

\subsubsection{Local Deformation Abnormalities}

\begin{figure*}[!t]
    \centering
    \begin{minipage}{.49\textwidth}
        \centering
        \centerline{\includegraphics[width=\columnwidth]{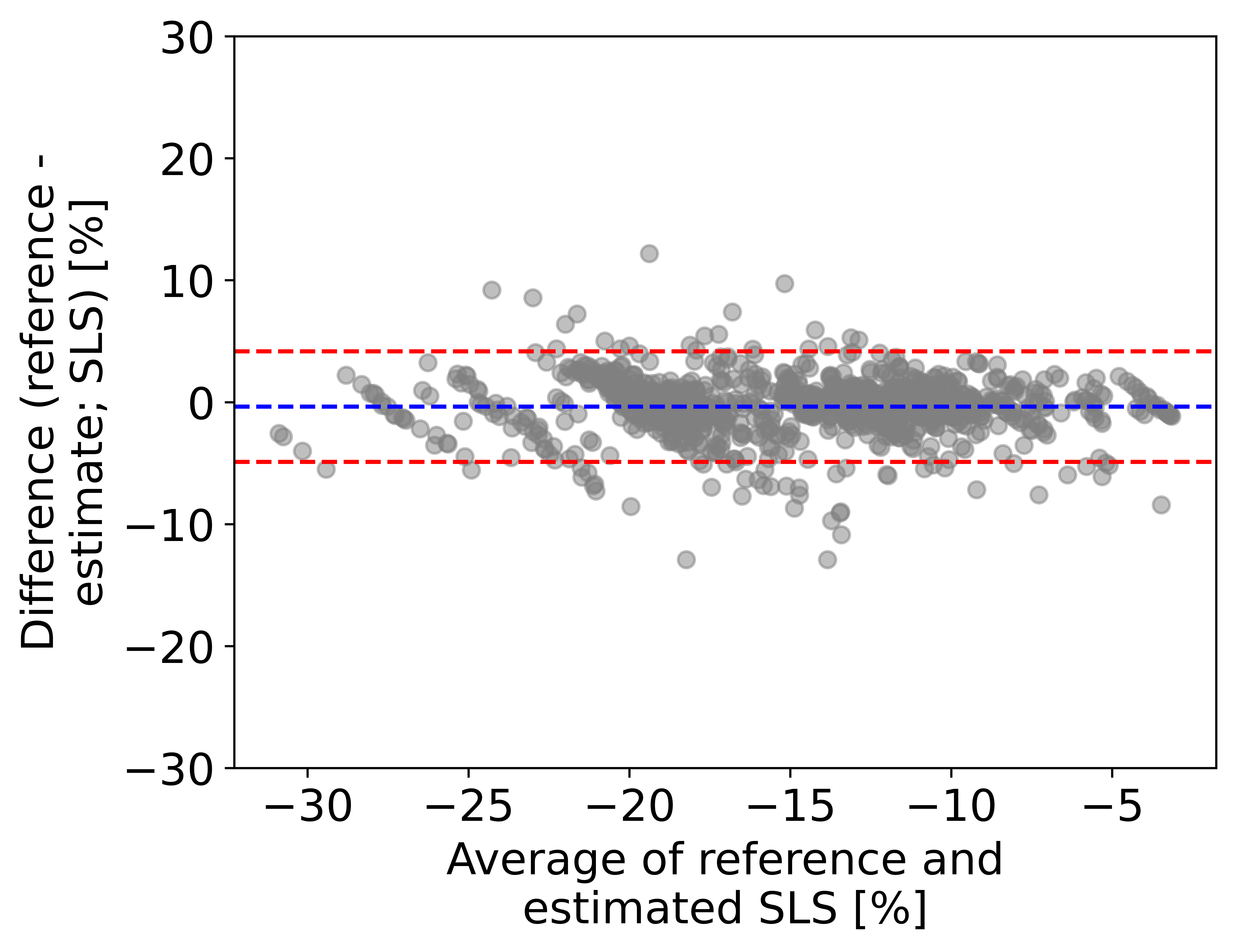}}
    \end{minipage}
    \begin{minipage}{.49\textwidth}
        \centering
        \centerline{\includegraphics[width=\columnwidth]{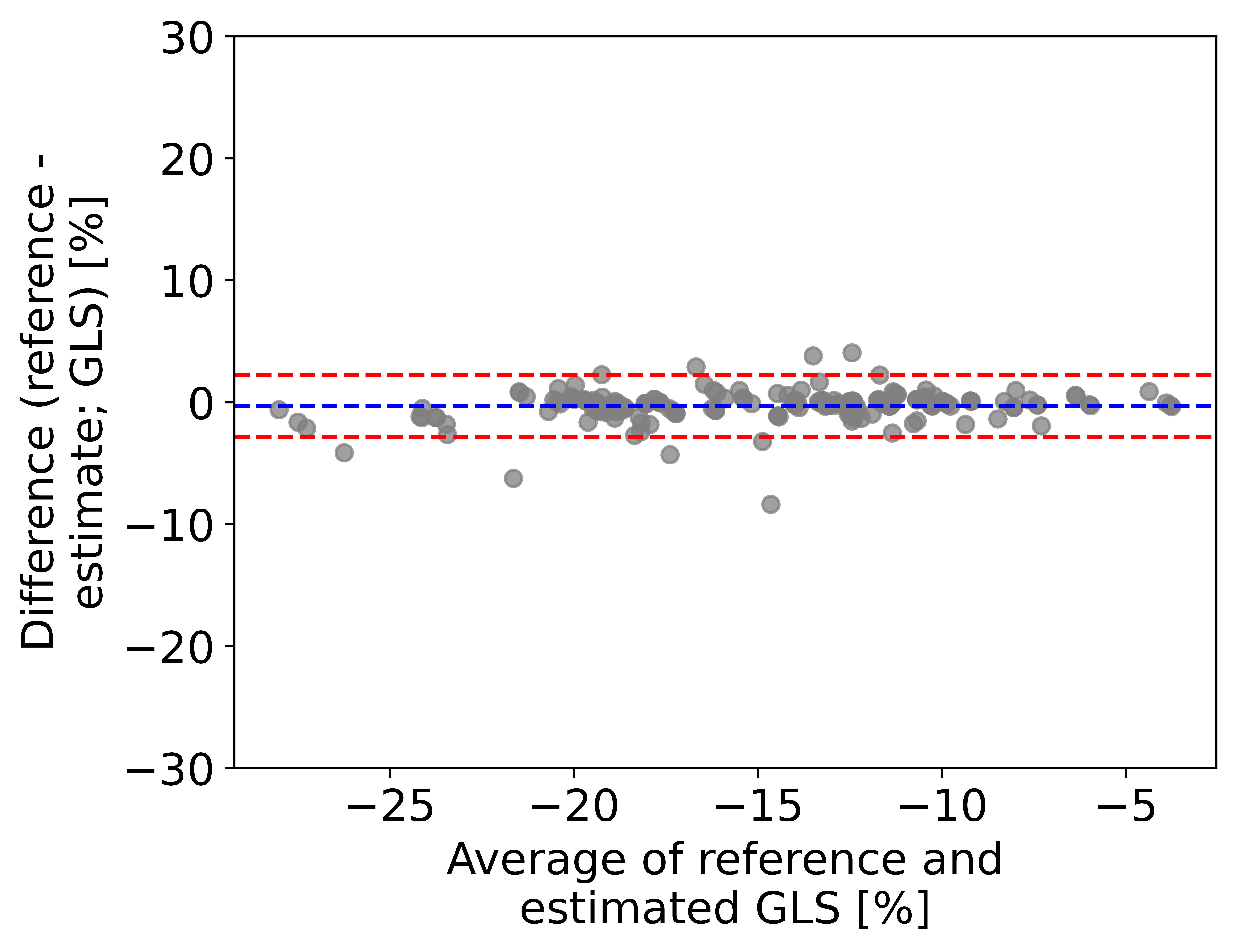}}
    \end{minipage}
    \caption{Bland–Altman plot comparing manual and automatically estimated SLS and GLS measures in synTEE data, including all four datasets (Table \ref{tab:datasets_ratio}). TeeTracker was used for automatic measurements. The blue line indicates the mean difference, and the red lines indicate the 95\% limits of agreement. SLS, segmental longitudinal strain.}
    \label{fig:ba_plot_syn_inf}
\end{figure*}

We extended our pipeline to simulate myocardial infarction, aiming to assess our model's ability to detect and adapt to localized hypokinetic dysfunction in cardiac muscle. This was done through an ablation study whose results were given in Table \ref{tab:synTEE_synt_inf_results}. Using the combined strategy, we trained our TeeTracker on the fourth dataset described in Table \ref{tab:datasets_ratio}, with and without a fifth dataset involving synthetic infarction. The model trained on the dataset incorporating synthetic infarction demonstrated remarkable generalization to regional variations in myocardial contraction, with a mean distance error of \mbox{0.37 $\pm$ 0.06 mm} across all segments and \mbox{0.36 $\pm$ 0.11 mm} on infarcted segments. Moreover, this model outperformed the one trained without synthetic infarction, which achieved a mean distance error of \mbox{0.58 $\pm$ 0.14 mm} across all segments. These results also hold for the clinical metrics, where TeeTracker trained on the combined dataset with synthetic infarcts achieved the best scores with a mean difference (95\% limits of agreement) of \mbox{0.14\% (-2.60\% to 2.88\%)} for SLS and a mean difference (95\% limits of agreement) of \mbox{0.11\% (-0.69\% to 0.91\%)} for GLS. The quality of these results was further validated through the \mbox{Bland-Altman} plots provided in Figure \ref{fig:ba_plot_syn_inf}. Finally, Figure \ref{fig:prediction_examples} shows the reference and segmental strain curves estimated by our TeeTracker models from a simulated sequence with an infarct region. Visual inspection of the curves reveals that our best model successfully identified and localized the infarcted segments, maintaining a coherent global contraction pattern while accurately reflecting regional dysfunction. These results demonstrate the ability of TeeTracker to detect and quantify myocardial abnormalities, thereby enhancing its clinical applicability for real-world TEE data.

\begin{table*}[tbph]
\caption{Comparison of performance in synTEE with synthetic infarction}
\label{tab:synTEE_synt_inf_results}
\begin{center}
\resizebox{\textwidth}{!}{%
\begin{tabular}{llllllccc}
\hline
\multicolumn{1}{c}{\multirow{2}{*}{Model}} & \multirow{2}{*}{Training scheme}                                                        & \multirow{2}{*}{Mean distance error {[}mm{]}} & \multirow{2}{*}{\begin{tabular}[c]{@{}l@{}}Mean distance error of\\ infarcted segment {[}mm{]}\end{tabular}} & \multicolumn{1}{c}{\multirow{2}{*}{Myocardial segments}} & \multicolumn{2}{c}{SLS {[}\%{]}}                                      & \multicolumn{2}{c}{GLS {[}\%{]}}                                               \\ \cline{6-9} 
\multicolumn{1}{c}{}                       &                                                                                         &                                               &                                                                                                              & \multicolumn{1}{c}{}                                     &                                   & SD                                & MD                                & SD                                         \\ \hline
\multirow{4}{*}{TeeTracker}                & \multirow{2}{*}{Combined}                                                               & \multirow{2}{*}{0.58$\pm$0.14}                 & \multirow{2}{*}{0.61$\pm$0.22}                                                                                & All                                                      & \multicolumn{1}{c}{0.11}          & \textit{3.20}                     & 0.14                              & \textit{0.88}                              \\
                                           &                                                                                         &                                               &                                                                                                              & Basal + mid                                              & 0.17                              & \multicolumn{1}{l}{\textit{3.32}} & 0.14                              & \textit{0.88}                              \\ \cline{2-9} 
                                           & \multirow{2}{*}{\begin{tabular}[c]{@{}l@{}}Combined\\ + synthetic infarct\end{tabular}} & \multirow{2}{*}{0.37$\pm$0.06}                 & \multirow{2}{*}{0.36$\pm$0.11}                                                                                            & All                                                      & 0.10                              & \multicolumn{1}{l}{\textit{1.65}} & \multicolumn{1}{l}{0.11}          & \multicolumn{1}{l}{\textit{0.41}}          \\
                                           &                                                                                         &                                               &                                                                                                              & Basal + mid                                              & \multicolumn{1}{c}{\textbf{0.14}} & \textit{\textbf{1.40}}            & \multicolumn{1}{l}{\textbf{0.11}} & \multicolumn{1}{l}{\textit{\textbf{0.41}}} \\ \hline
\end{tabular}%
}%
\end{center}
\hfill \break
{Comparison of performance at motion estimation between SOTA methods in synTEE with synthetic infarction. Motion estimation was evaluated directly by comparing the mean Euclidean distance between the reference mesh and the myocardial mesh throughout the sequence. In addition, we evaluated the agreement of measures between estimated and reference SLS and GLS measures in synTEE with synthetic dysfunction in myocardial contraction. SLS, segmental longitudinal strain. GLS, global longitudinal strain. MD, mean difference. SD, standard deviation.}
\end{table*}

\begin{figure*}[tbph] 
    \centering
    \begin{minipage}{.24\textwidth}
        \centering
        \includegraphics[width=\textwidth]{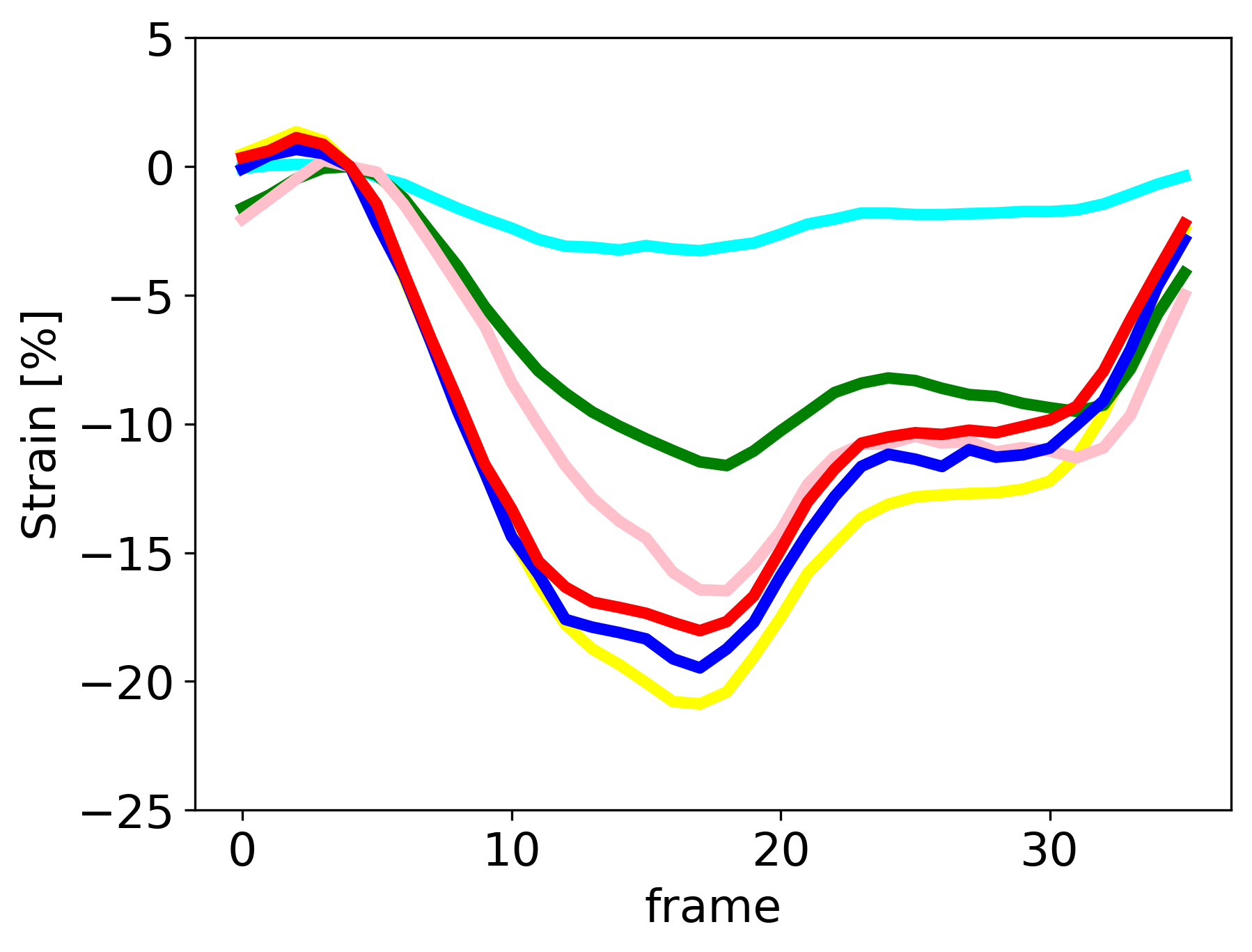}
        (a)
    \end{minipage}
    \begin{minipage}{.24\textwidth}
        \centering
        \includegraphics[width=\textwidth]{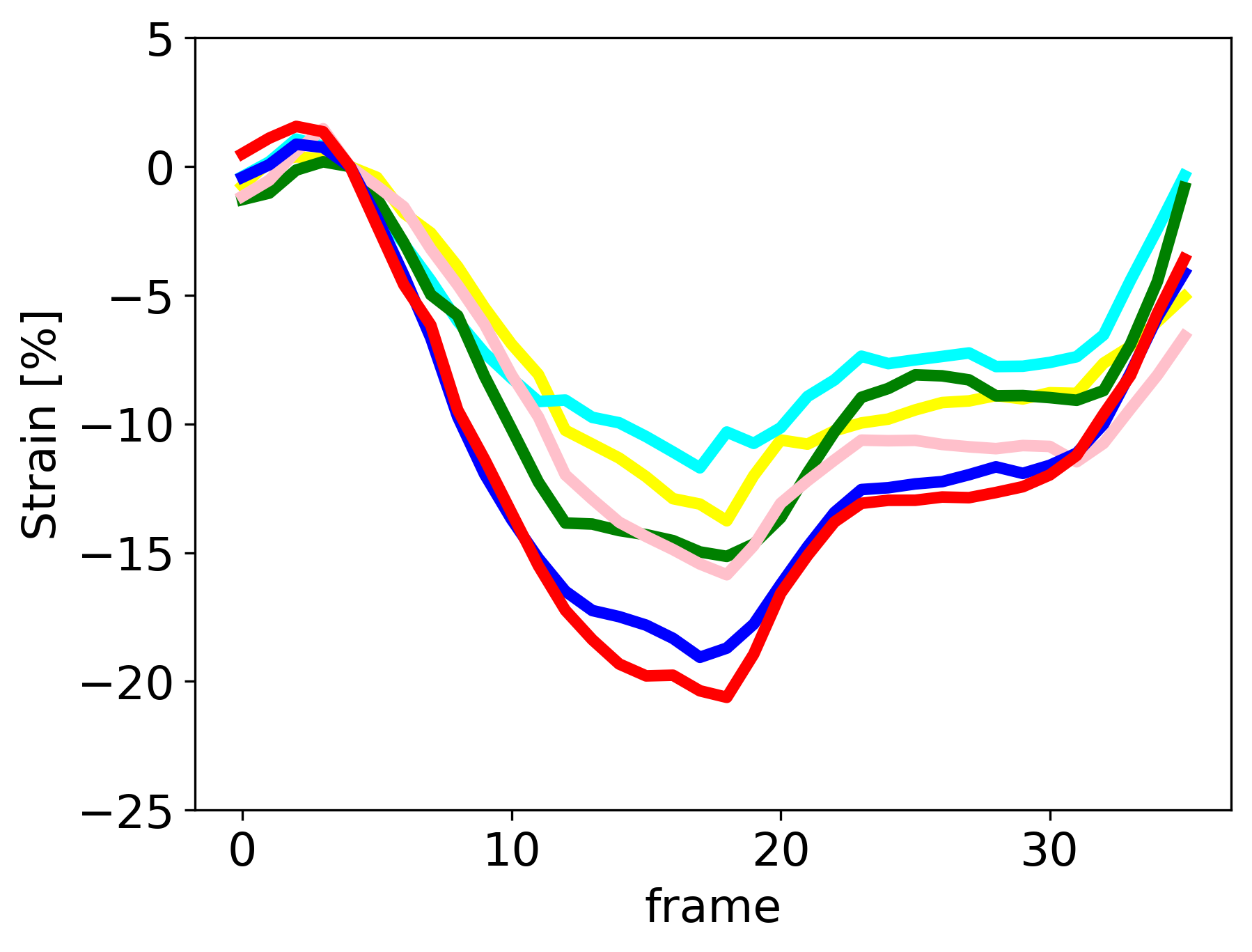}
        (b)
    \end{minipage}
    \begin{minipage}{.24\textwidth}
        \centering
        \includegraphics[width=\textwidth]{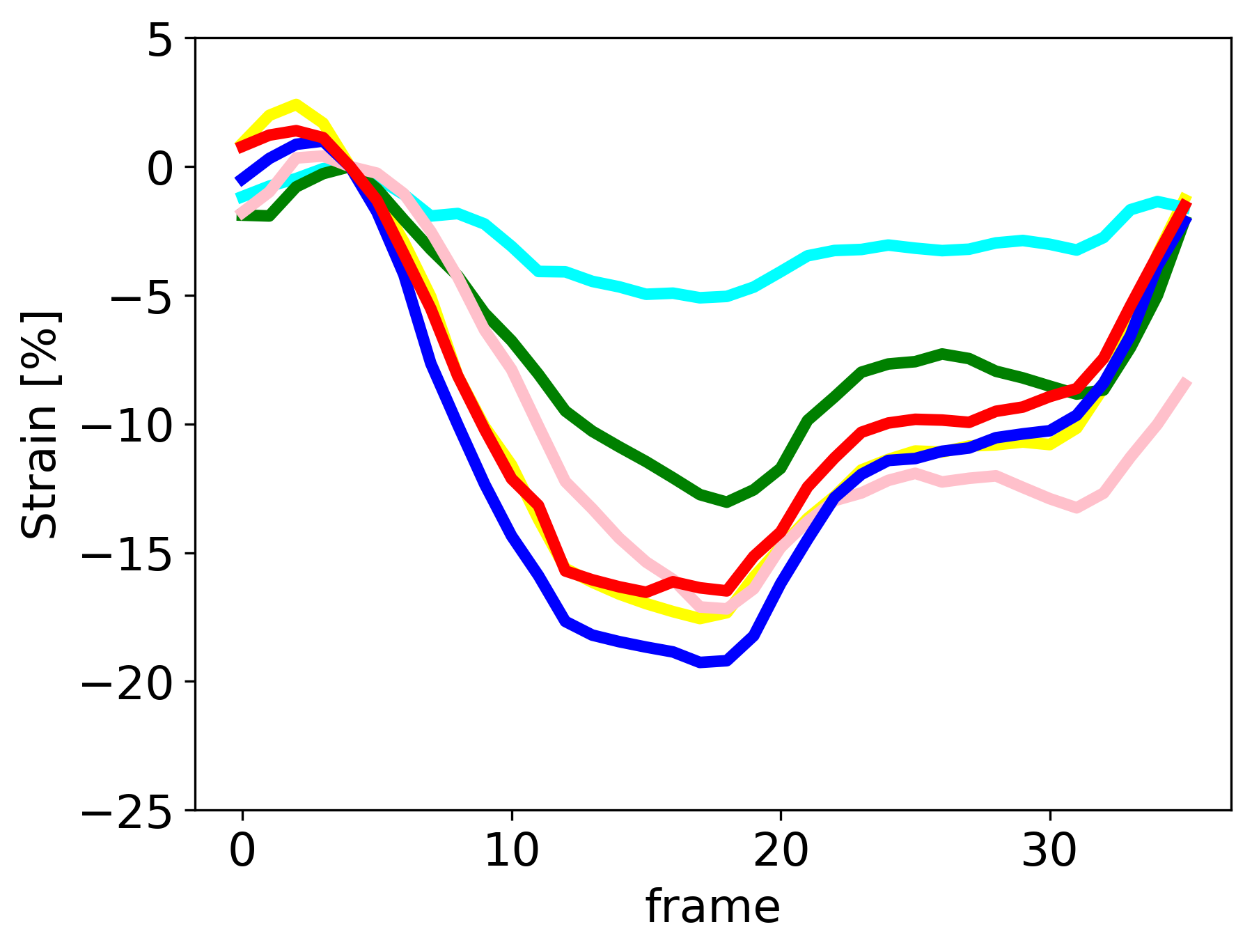}
        (c)
    \end{minipage}
    \begin{minipage}{.24\textwidth}
        \centering
        \includegraphics[width=\textwidth]{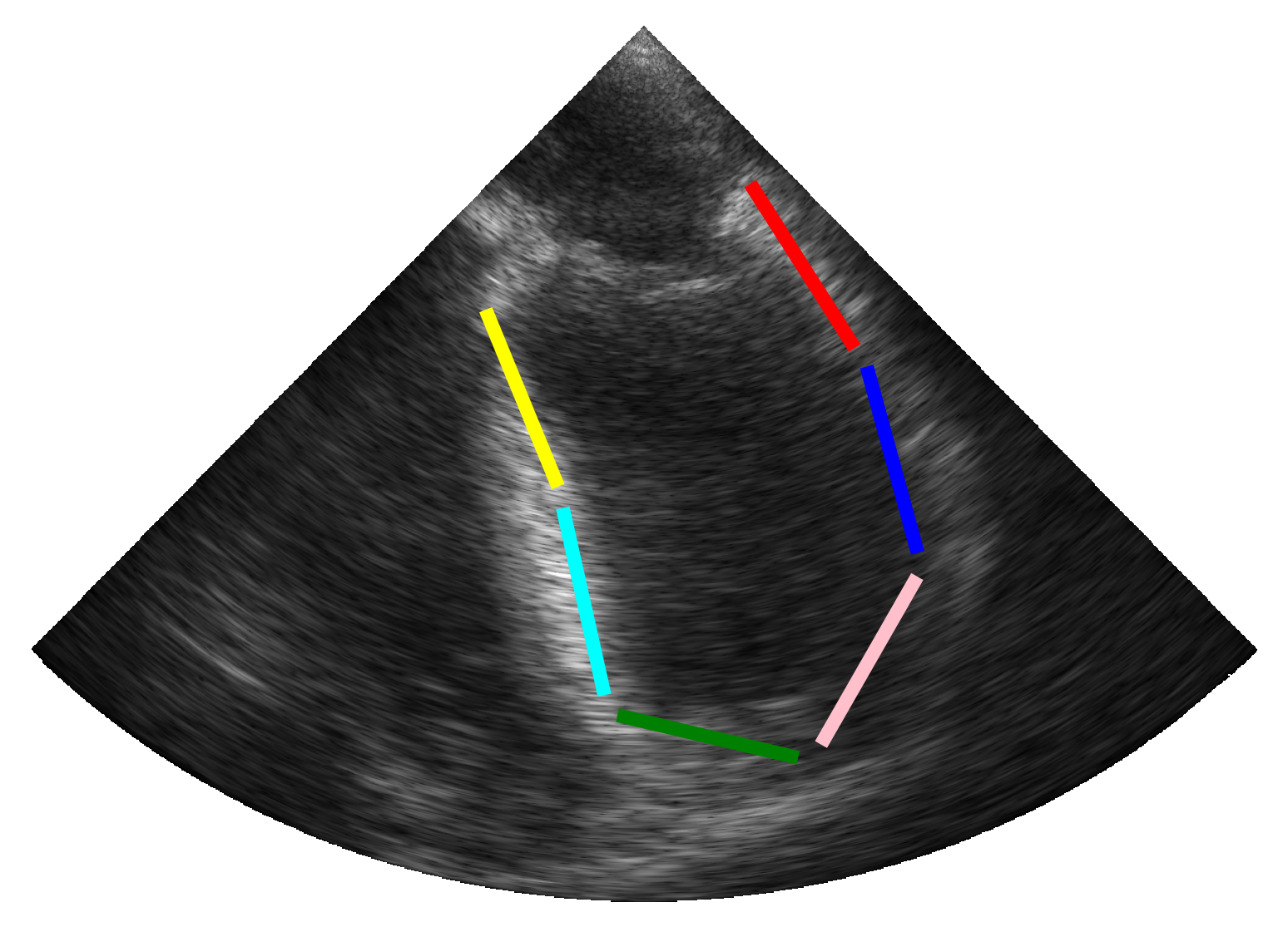}
        (d)
    \end{minipage}
    \caption{Examples of SLS curves obtained in synTEE test data by (a) ground truth references with a synthetic infarction in the mid septal segment (cyan); (b) TeeTracker trained in a combined scheme; (c) TeeTracker trained in a combined scheme with synthetic infarcted segments (d) visualization of the colormap for myocardial segments.}
    \label{fig:prediction_examples}
\end{figure*}

\subsection{Clinical Data}

\begin{figure*}[tbph]
    \centering
    \begin{minipage}{.49\textwidth}
        \centering
        \centerline{\includegraphics[width=\columnwidth]{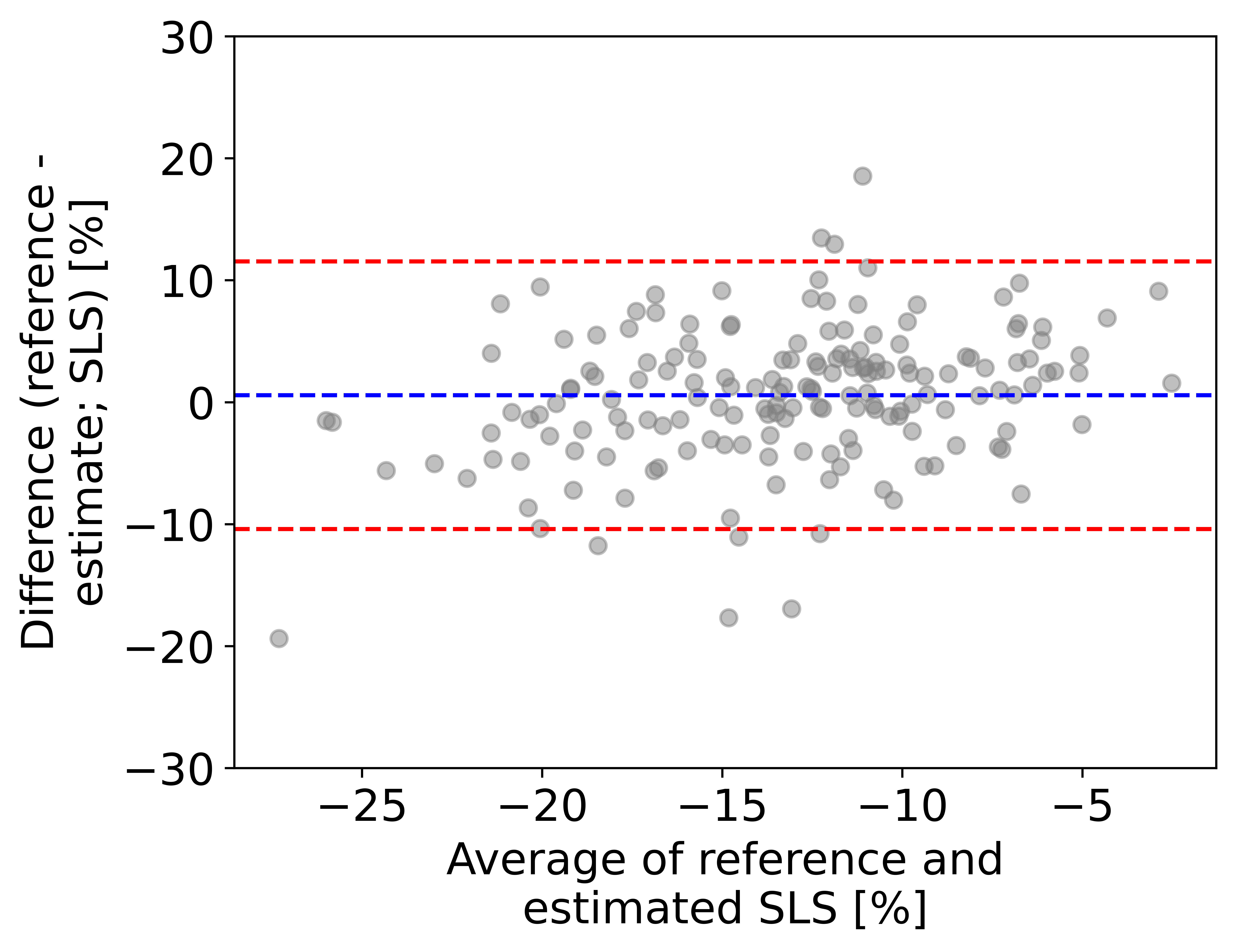}}
        (A)
    \end{minipage}
    \begin{minipage}{.49\textwidth}
        \centering
        \centerline{\includegraphics[width=\columnwidth]{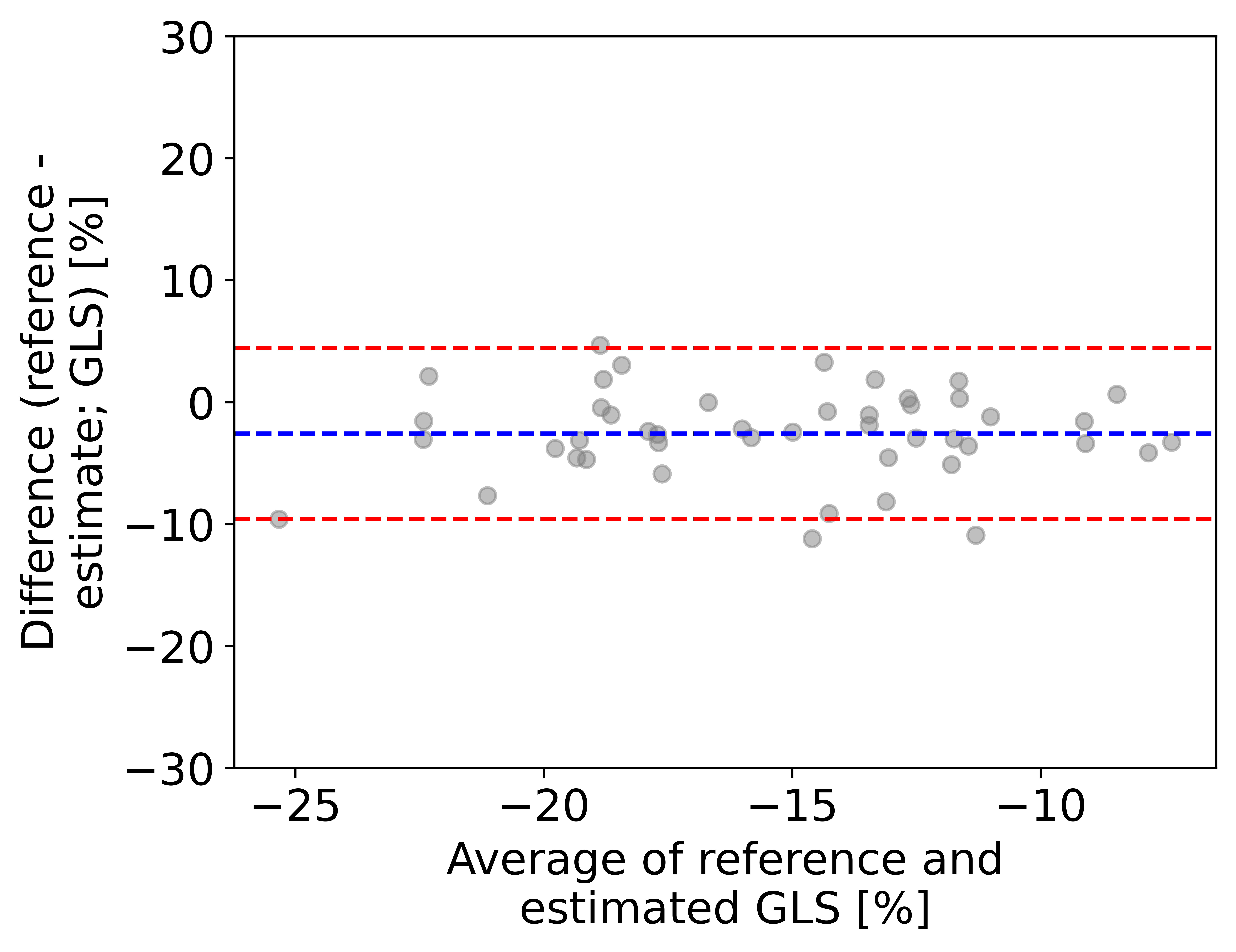}}
        (B)
    \end{minipage}
    \caption{Bland–Altman plot comparing manual and automatically estimated SLS and GLS measures in real TEE. TeeTracker is used for automatic measurements. The blue line indicates the mean difference and the red lines indicate the 95\% limits of agreement. SLS, segmental longitudinal strain.}
    \label{fig:ba_plot_tee}
\end{figure*}

Table \ref{tab:TEE_results} presents the clinical scores obtained by TeeFlow and TeeTracker using real datasets and employing a combined scheme with and without synthetic infarcts (for TeeTracker only). As in the case of synthetic data, TeeTracker trained on the combined dataset with synthetic infarcts achieved the best results for the estimation of GLS and SLS across all segments, with mean differences of \mbox{-2.36\% (-8.36\% to 3.52\%)} and \mbox{1.09\% (-8.90\% to 11.09\%)}, respectively. The exclusion of apical segments in the estimation of SLS yielded more contrasted results, with TeeFlow achieving the best mean difference of -0.14\%, while TeeTracker exhibited the lowest standard deviation at 5.10\%. While TeeTracker trained with synthetic infarcts achieved most of the best results, its scores remained close to those of the same model trained without synthetic infarcts. We therefore present in Table \ref{tab:TEE_results_syn_inf} the scores of these two models computed only on hypokinetic segments where the reference strain was below 5\%. The results showed that incorporating synthetic data with local deformation abnormalities improved TeeTracker's accuracy and precision in estimating SLS on segments with low reference strain, indicating enhanced ability to detect myocardial infarction. Finally, we displayed in Figure \ref{fig:ba_plot_tee} the Bland-Altman plot comparing manual and automatically estimated SLS and GLS in real TEE. Interestingly, these results show that the performance in estimating GLS remains consistent when compared to results obtained from synthetic datasets. However, there was a noticeable decline in SLS estimation performance, with a ratio of around 4 compared to the scores obtained from the synthetic dataset. This indicates that there were still efforts needed in synthetic datasets to maintain TeeTracker's performance on local strain estimates when moving to real data.



\begin{table*}[tbph]
\caption{Comparison of performance in TEE}
\label{tab:TEE_results}
\begin{center}
\resizebox{.7\textwidth}{!}{%
\begin{tabular}{lllcccc}
\hline
\multicolumn{1}{c}{\multirow{2}{*}{Model}} & \multirow{2}{*}{Training scheme}                                       & \multicolumn{1}{c}{\multirow{2}{*}{Myocardial segments}} & \multicolumn{2}{c}{SLS {[}\%{]}}                             & \multicolumn{2}{c}{GLS {[}\%{]}}                                                       \\ \cline{4-7} 
\multicolumn{1}{c}{}                       &                                                                        & \multicolumn{1}{c}{}                                     & MD                                & SD                       & MD                                         & SD                                        \\ \hline
TeeFlow                                    & Combined                                                               & All                                                      & -2.67                             & \textit{7.42}            & \multirow{2}{*}{-2.69}                     & \multirow{2}{*}{\textit{3.69}}            \\
                                           &                                                                        & Basal + mid                                              & -0.14                             & \textit{6.21}            &                                            &                                           \\ \hline
TeeTracker                                 & Combined                                                               & All                                                      & \multicolumn{1}{l}{-2.33}         & \multicolumn{1}{l}{7.34} & \multicolumn{1}{l}{\multirow{2}{*}{-2.56}} & \multicolumn{1}{l}{\multirow{2}{*}{3.57}} \\
                                           &                                                                        & Basal + mid                                              & \multicolumn{1}{l}{\textbf{0.58}} & \multicolumn{1}{l}{5.60} & \multicolumn{1}{l}{}                       & \multicolumn{1}{l}{}                      \\ \hline
TeeTracker                                 & \begin{tabular}[c]{@{}l@{}}Combined\\ + synthetic infarct\end{tabular} & All                                                      & -2.28                             & \textit{7.37}            & \multirow{2}{*}{\textbf{-2.36}}            & \multirow{2}{*}{\textit{\textbf{3.06}}}   \\
                                           &                                                                        & Basal + mid                                              & 1.09                              & \textit{\textbf{5.10}}   &                                            &                                           \\ \hline
\end{tabular}
}%
\end{center}
\hfill \break
\hfill \break
\hfill \break
{Comparison of performance at motion estimation between SOTA methods in real TEE. We evaluated the agreement of measures between estimated and reference SLS and GLS measures in TEE. SLS, segmental longitudinal strain. GLS, global longitudinal strain. MD, mean difference. SD, standard deviation.}
\end{table*}

\begin{table*}[tbph]
\caption{Comparison of performance in TEE}
\label{tab:TEE_results_syn_inf}
\begin{center}
\resizebox{.6\textwidth}{!}{%
\begin{tabular}{lllll}
\hline
\multicolumn{1}{c}{\multirow{2}{*}{Model}} & \multirow{2}{*}{Training scheme}                                       & \multicolumn{1}{c}{\multirow{2}{*}{Myocardial segments}} & \multicolumn{2}{c}{SLS {[}\%{]}}                \\ \cline{4-5} 
\multicolumn{1}{c}{}                       &                                                                        & \multicolumn{1}{c}{}                                     & \multicolumn{1}{c}{MD} & \multicolumn{1}{c}{SD} \\ \hline
TeeTracker                                 & Combined                                                               & Infarct                                                  & 6.20                   & 4.13                   \\ \hline
TeeTracker                                 & \begin{tabular}[c]{@{}l@{}}Combined\\ + synthetic infarct\end{tabular} & Infarct                                                  & 4.88                   & 2.94                   \\ \hline
\end{tabular}
}%
\end{center}
\hfill \break
\hfill \break
\hfill \break
{Comparison of performance at motion estimation in myocardial segments with low reference SLS in real TEE. We evaluated the agreement of measures between estimated and reference SLS measures in TEE. We defined infarction as myocardial segments with a reference SLS of less than 5\%. SLS, segmental longitudinal strain. GLS, global longitudinal strain. MD, mean difference. SD, standard deviation.}
\end{table*}

\section{Discussion}

The integration of AI-based tracking in echocardiography through TeeTracker presents automated measuring of regional LV function. Our findings illustrate that DL can effectively interpret complex cardiovascular dynamics captured through TEE, aligning closely with both synthetic benchmarks and clinical expectations. Training and evaluating tracking methods in TEE were feasible thanks to the generation of synthetic echocardiographic sequences with ground truth motion references. TeeTracker provided precise motion estimates in synthetic data, with a mean distance error of 0.65 mm across the various datasets. Furthermore, TeeTracker demonstrated the ability to generalize to real data for both GLS and SLS estimation, achieving a mean difference (95\% limits of agreement) of \mbox{1.09\% (-8.90\% to 11.09\%)} for SLS.


\subsection{Synthetic Data}

Evaluations using synthetic data demonstrated the accuracy and precision of our DL models, TeeFlow and TeeTracker, in estimating myocardial motion, GLS, and SLS from TEE sequences. Notably, TeeTracker outperformed TeeFlow, exhibiting a significantly lower mean distance error across various test sets. This emphasizes the strength of the point trajectory estimation method in managing the complexities of myocardial motion. Including various levels of speckle decorrelation in the synthetic datasets was crucial for enhancing the model's adaptability to real-world conditions, highlighting the importance of realistic data simulation in training deep learning models.

\added{While real echocardiographic data might better capture clinical imaging complexity, this approach presents significant challenges: (1) obtaining ground truth motion fields for TEE is virtually impossible without invasive markers; (2) expert annotations introduce inter-observer variability; and (3) acquiring adequately sized datasets with diverse pathologies requires extensive annotation efforts.}

\subsection{Influence of the Training Procedure}

Our investigation of various training schemes revealed that a combined training approach yielded the best overall performance. This approach allowed the model to effectively learn fundamental contraction patterns while adapting to varying degrees of speckle decorrelation. The significant performance improvement after fine-tuning, especially with the combined scheme, validates our hypothesis that exposure to a broad spectrum of decorrelation scenarios enhances model robustness. Moreover, the ability of TeeTracker to maintain accuracy across different synthetic datasets indicates its potential for generalizability in clinical settings.

\subsubsection{Estimation of Strain Measures}

The performance of the \emph{autoStrain} pipeline in estimating SLS and GLS further reinforces the efficacy of our approach. The high agreement between estimated and reference measures, particularly for basal and mid segments, demonstrates the clinical relevance of our models. The level of precision achieved was crucial for the reliable monitoring of cardiac function in perioperative and critical care settings. The poor performance in motion estimation of apical segments was anticipated, due to severely foreshortened TEE images with noticeable out-of-plane movement. In addition, the apical segments were distant from the ultrasound probe, resulting in higher decorrelation due to attenuation.

\subsubsection{Local Deformation Abnormalities}

Incorporating synthetic infarction into the training datasets significantly improved the model's ability to detect and quantify reduced regional myocardial contraction. This enhanced sensitivity to regional variations underscores the potential of our approach for identifying and localizing myocardial infarctions. Such a characteristic was true also for real TEE, with a better accuracy and precision in SLS estimation of segments with low references. The resulting strain measures were highly consistent with reference values, confirming the clinical applicability of our method in detecting localized cardiac abnormalities.

\subsection{Clinical Data}

The translation of TeeTracker's performance from synthetic datasets to clinical data was promising. This consistency highlights the robustness of the model and its potential for integration into clinical workflows. \added{Our models were trained exclusively on synthetic data and directly applied to real TEE patients without fine-tuning. This approach demonstrates the robustness of our methods and the realism of our simulation pipeline.} The ability of TeeTracker to provide reliable SLS measurements in real-world clinical settings suggests that our AI-driven approach can meet current clinical standards, paving the way for its adoption in routine cardiac diagnostics. These results were particularly encouraging since the real clinical data provided was substantially corrupted by noise and out-of-plane movement due to foreshortening, as the images were acquired with a passive probe placement in the esophagus. To our knowledge, no other studies have reported on the automatic estimation of SLS.

The GLS estimation by TeeTracker showed promising results when compared to those reported in TTE. In studies with larger and more comprehensive datasets, Østvik \mbox{et al.} \cite{ostvik_strain} reported a mean difference (95\% limits of agreement) in GLS estimation relative to clinical standards of \mbox{-0.71\% (-3.90\% to 2.48\%)}. Similarly, Evain \mbox{et al.} \cite{creatis_synTTE_strain} found a mean difference of \mbox{2.50\% (-1.61\% to 6.62\%)}, while Azad \mbox{et al.} \cite{azad2024echo} documented a mean difference of \mbox{-0.13\% (-3.62\% to 3.36\%)}. The TTE data in the referenced studies likely included less foreshortened and noisy images compared to our TEE dataset, as our images were acquired with passive probe placement to simulate a monitoring use case of the \emph{autoStrain} system.

\subsection{Limitations and Future Work}

Challenges remain, particularly in managing the inherent variability of clinical data and ensuring the generalization of AI systems across different patient populations. \added{Our clinical validation dataset of 16 patients is relatively small, limiting statistical power and pathology diversity. While our synthetic training approach mitigates overfitting concerns, future work should include larger clinical datasets to further validate generalizability.} However, the consistency of our model's performance from synthetic training to real clinical application suggests that, with further refinement, these tools can become a standard part of cardiac diagnostics, enhancing the precision and efficiency of patient care.

Although leveraging synthetic data in training our AI models, TeeFlow and TeeTracker, has enabled the circumvention of common data scarcity issues and has provided a controlled environment to refine our algorithms, there were inherent limitations that must be acknowledged. 
Primarily, synthetic datasets, although sophisticatedly designed to mimic real physiological conditions, may not fully capture the complex variability seen in actual patient data. For instance, factors such as patient-specific myocardial responses to pathology, variabilities in echocardiographic image quality, complex myocardial contraction patterns, and interpatient anatomical differences were challenging to simulate comprehensively. These limitations could potentially impact the generalizability and clinical applicability of our findings. The models trained on synthetic data performed well under controlled test conditions; however, their performance in real-world scenarios may exhibit variability not captured in this study. \added{Our study was limited to data from a single vendor (GE). Future work should evaluate robustness across multiple vendors, as ultrasound characteristics vary between manufacturers. Cross-vendor validation could also be achieved through transfer learning or domain adaptation techniques.} Future research should aim to incrementally integrate real patient datasets into the training process, enhancing the robustness and adaptability of the models to handle the broad spectrum of clinical presentations encountered in practice. This step was vital for ensuring that our AI-driven methodologies achieve consistent reliability across diverse clinical settings, ultimately supporting their adoption into routine clinical use.

Future research should focus on expanding the training datasets to include a broader array of cardiac conditions and patient demographics to enhance the models' generalization capabilities. To validate the methods ability to detect regional dysfunction, a bigger test dataset of real patients with infarcted segments was needed. Additionally, integrating these AI tools into real-time clinical workflows could significantly decrease the time needed for diagnostic assessments, a critical factor in acute cardiac care settings.

\section{Conclusion}

The study confirmed the effectiveness of using AI-driven tools for estimating myocardial motion, GLS and SLS in TEE. TeeFlow and TeeTracker not only demonstrated high accuracy in synthetic validations but also showed promising clinical applicability. This suggests that AI can play a crucial role in advancing cardiac diagnostic methodologies, potentially improving patient outcomes through more precise and timely assessments. Our next steps will involve further clinical validations and system refinements to ensure that these tools can be seamlessly integrated into everyday clinical practice, marking a significant step forward in the monitoring of regional LV function of critically ill patients.

\appendices

\section*{Acknowledgment}

We extend our gratitude to the Clinic of Cardiology and the Clinic of Anesthesia and Intensive Care at St. Olavs University Hospital that provided clinical data and expertise, significantly contributing to the research's success. Additionally, we thank the technical team at CREATIS, INSA, Lyon, France for developing and maintaining the method for in silico modeling of myocardial motion used in this study.

\bibliographystyle{abbrv}
\bibliography{references}

\begin{IEEEbiography}[{\includegraphics[width=1in,height=1.25in,clip,keepaspectratio]{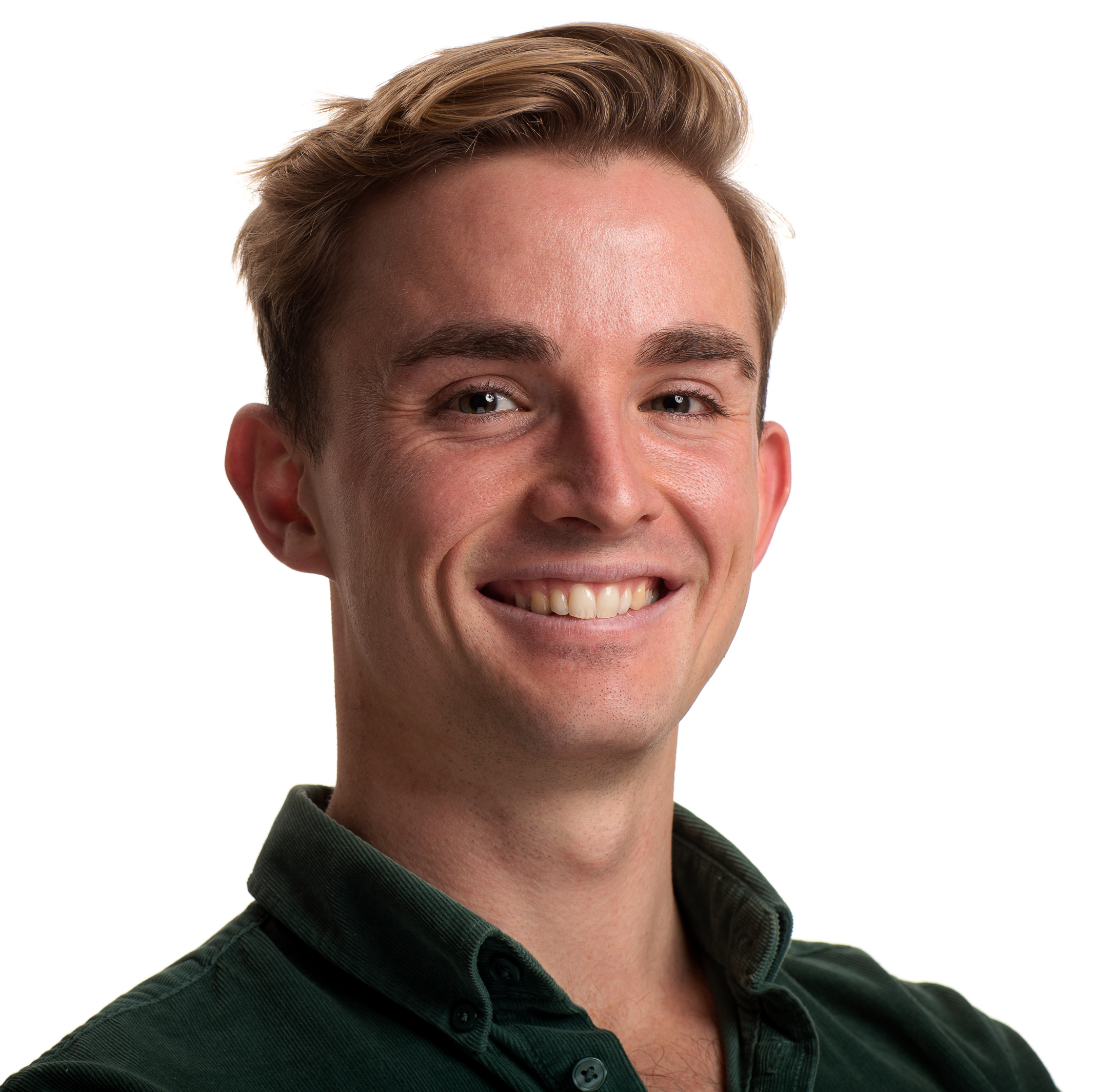}}]{Anders Austlid Taskén} was born in Bærum, Norway in 1996. He received M.Sc. in Cybernetics and Robotics at the University of Science and Technology (NTNU) in 2021. He started his doctoral degree in 2021 and is currently pursuing the Ph.D. at the Department of Computer Science, Faculty of Information Technology and Electrical Engineering, Norwegian University of Science and Technology. His research interests include computerized artificial intelligence on cardiac ultrasound, with a focus on peri- and post-operative automatic monitoring a left ventricular function in critically ill patients. In 2023, he completed a research stay abroad at CREATIS, INSA, Lyon and initiated an international collaboration with University of Sherbrooke, INSA and NTNU.
\end{IEEEbiography}

\end{document}